\documentclass[lettersize,journal]{IEEEtran}
\usepackage{amsmath,amsfonts}
\usepackage{algorithmic}
\usepackage{algorithm}
\usepackage{array}
\usepackage[caption=false,font=normalsize,labelfont=sf,textfont=sf]{subfig}
\usepackage{textcomp}
\usepackage{stfloats}
\usepackage{url}
\usepackage{verbatim}
\usepackage{graphicx}
\usepackage{cite}

\usepackage[utf8]{inputenc} 
\usepackage[T1]{fontenc}    
\usepackage{hyperref}       
\usepackage{url}            
\usepackage{booktabs}       
\usepackage{amsfonts}       
\usepackage{nicefrac}       
\usepackage{microtype}      
\usepackage{xcolor}         
\usepackage{enumitem}
\usepackage{multirow}
\usepackage{amsthm}
\usepackage{amsmath}
\usepackage{amsfonts}
\usepackage{color}
\usepackage{pifont}

\begin{document}

\title{Comprehensive Graph Gradual Pruning  \\ for Sparse Training in Graph Neural Networks}

\author{Chuang Liu\thanks{This work has been done when Chuang Liu worked as an intern at JD Explore Academy.}, Xueqi Ma, Yibing Zhan,~\IEEEmembership{Member,~IEEE,} Liang Ding,~\IEEEmembership{Member,~IEEE,} Dapeng Tao,~\IEEEmembership{Member,~IEEE,} \\ Bo Du,~\IEEEmembership{Senior Member,~IEEE,} Wenbin Hu, and Danilo Mandic,~\IEEEmembership{Fellow,~IEEE}
\thanks{Chuang Liu, Bo Du, and Wenbin Hu are with the School of Computer Science, Wuhan University, Hubei, China (E-mail: chuangliu@whu.edu.cn; dubo@whu.edu.cn; hwb@whu.edu.cn).}
\thanks{Xueqi Ma and Danilo Mandic are with Department of Electrical Engineering, Imperial College, London, U.K. (E-mail: xueqima@s.upc.edu.cn; d.mandic@imperial.ac.uk). }

\thanks{Yinbing Zhan and Liang Ding are algorithm scientists at the JD Explore Academy, Beijing, China (E-mail: zhanyibing@jd.com; dingliang1@jd.com).}

\thanks{Dapeng Tao is with the School of Information Science and Engineering, Yunnan University, Kunming, China (E-mail: dptao@ynu.edu.cn).}

}

\markboth{Journal of \LaTeX\ Class Files,~Vol.~14, No.~8, August~2021}%
{Shell \MakeLowercase{\textit{et al.}}: A Sample Article Using IEEEtran.cls for IEEE Journals}


\maketitle

\begin{abstract}

Graph Neural Networks (GNNs) tend to suffer from high computation costs due to the exponentially increasing scale of graph data and the number of model parameters, which restricts their utility in practical applications. To this end, some recent works focus on sparsifying GNNs (including graph structures and model parameters) with the lottery ticket hypothesis (LTH) to reduce inference costs while maintaining performance levels. However, the LTH-based methods suffer from two major drawbacks: \textbf{1)} they require exhaustive and iterative training of dense models, resulting in an extremely large training computation cost, and  \textbf{2)} they only trim graph structures and model parameters but ignore the node feature dimension, where significant redundancy exists. To overcome the above limitations, we propose a \textsc{\bf c}omprehensive graph \textsc{\bf g}radual \textsc{\bf p}runing framework termed \textsc{\bf CGP}. This is achieved by designing a during-training graph pruning paradigm to dynamically prune GNNs within one training process. Unlike LTH-based methods, the proposed CGP approach requires no re-training, which significantly reduces the computation costs. Furthermore, we design a co-sparsifying strategy to comprehensively trim all three core elements of GNNs: graph structures, node features, and model parameters. Meanwhile, aiming at refining the pruning operation, we introduce a regrowth process into our CGP framework, in order to re-establish the pruned but important connections. The proposed CGP is evaluated by using a node classification task across 6 GNN architectures, including shallow models (GCN and GAT), shallow-but-deep-propagation models (SGC and APPNP), and deep models (GCNII and ResGCN), on a total of 14 real-world graph datasets, including large-scale graph datasets from the challenging Open Graph Benchmark. Experiments reveal that our proposed strategy greatly improves both training and inference efficiency while matching or even exceeding the accuracy of existing methods.\footnote{Codes will be released after the publication.}

\end{abstract}

\begin{IEEEkeywords}
Graph neural networks (GNNs), sparse training, gradual pruning, homophily and heterophily, node classification.
\end{IEEEkeywords}

\section{Introduction}
\label{sec:introduction}

Graph Neural Networks (GNNs)~\cite{gcn,graphsage,gat,gin} have achieved notable success in a variety of applications~\cite{tnnls-multi-gnn,tnnls-hawk,tnnls-gig,tnnls-graph-drawing} and have consequently become a rapidly growing area of research~\cite{tnnls-gnn-survey,survey-1,survey-2,survey-3}. Despite success, with the size of the graph data~\cite{ogb-dataset} and the complexity of the model structure increasing~\cite{deepgcns,deepgcns-tpami,gcnii}, GNNs exhibit experientially computation costs, during both training and inference~\cite{acceleration-survey}. This has been prohibitive to their deployment in resource-constrained or time-sensitive applications. For example, a 2-layer GCN model~\cite{gcn} with 32 hidden units requires approximately 19 GFLOPs (flops: floating point operations) to process the Reddit graph~\cite{degree-quant,gebt}, twice as much as the requirements of the ResNet50 model~\cite{resnet} on ImageNet. If the number of layers of GCN model rises to 64, the model requires approximately 73 GFLOPs. Such enormous computation costs of GNNs are primarily attributable to \underline{three aspects}: \textbf{1)} the large-scale adjacency matrix in real-world datasets (e.g., millions of edges in OGB datasets~\cite{ogb-dataset}), \textbf{2)} the high dimension node feature vectors (e.g., each node in the Citeseer graph has 3,703 features), and \textbf{3)} the sheer number of model parameters (e.g., a 64-layer GCNII~\cite{gcnii} with 64 hidden units contains about 262,144 parameters).

To reduce the enormous computation costs, several approaches (i.e., UGS~\cite{ugs}, GEBT~\cite{gebt}, and ICPG~\cite{icpg}) have been developed on the basis of the lottery ticket hypothesis (LTH)~\cite{lth}. Specifically, these methods propose to generalize the popular LTH to GNNs in order to find a subnetwork and a subgraph that best preserve the GNN performance in a sparse manner. The LTH-based approaches usually adopt an iterative cycle (train-then-sparsify), i.e., 1) train-dense, 2) prune, 3) re-train, focusing on sparsifying edges and model weights for the principal purpose of reducing inference cost.

\begin{figure*}[!t] 
\setlength{\abovecaptionskip}{-0.3cm}   
\begin{center}
\includegraphics[width=1.0\linewidth]{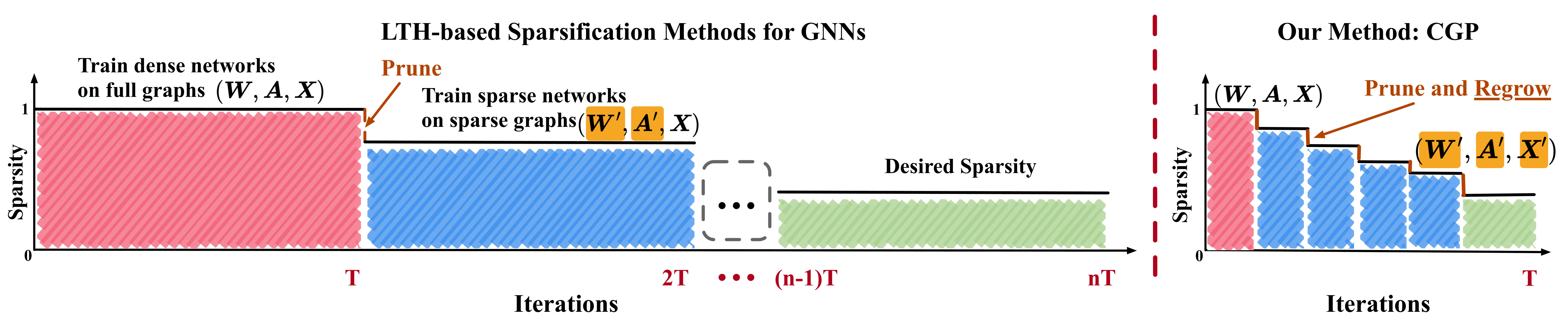}
\end{center}
\caption{The proposed CGP method vs. LTH-based sparsification methods. \textit{Left.} LTH-based pruning methods. \textit{Right.} Our proposed method, where $T$ refers to a whole training process when the model is trained to converge. The detailed definitions of $\boldsymbol{W}$, $\boldsymbol{A}$ and $\boldsymbol{X}$, whose pruned values are  $\boldsymbol{W}^{\prime}$, $\boldsymbol{A}^{\prime}$ and $\boldsymbol{X}^{\prime}$ respectively, can be found in Section~\ref{sec:preliminary}. }
\label{fig:compare}
\end{figure*}

Despite their good performance, the above LTH-based sparsification methods are still unsatisfactory due to: \textbf{1) Enormous Training Costs.} Existing methods necessitate training dense models exhaustively and iteratively (up to 20\textit{×}), which observably increases the training time (See the \textit{Left} part of Figure~\ref{fig:compare}).  Additionally, these methods cannot be implemented on low-capacity devices since the dense models must be stored in memory at the outset of the operation. \textbf{2) Redundancy of Node Features.} Existing techniques only co-sparsify the input graph structure (i.e., adjacency matrix) and model parameters, thereby neglecting the benefits of pruning the high-dimension node feature, which is a fundamental but redundant element in most GNNs.

To solve the aforementioned difficulties, we introduce a comprehensive graph gradual pruning framework, termed CGP, a novel and efficient sparse training framework for GNNs, whereby \textbf{1)} CGP employs a during-training graph pruning paradigm that enables the discovery of high-performing sparse GNNs in a single training procedure (See the \textit{Right} part of Figure~\ref{fig:compare}), as opposed to the LTH iterative pruning and retraining. \textbf{2)} Then, we leverage the unique structures of GNNs to comprehensively exploit GNN-specific pruning strategies, containing graph structures, node features, and model parameters, to sufficiently boost the training and inference efficiency of GNNs. \textbf{3)} Furthermore, we incorporate regrowth schemes into our framework to recover the "mistakenly" pruned connections (i.e., the significant connections that are pruned during the pruning process). In addition, CGP can support sparse-to-sparse training, which enables sparse initializations and maintains sparse values throughout training, thus further reducing the memory cost and accelerating the overall training process, hence facilitating the development of GNNs in real-time or source-constrained applications.

To evaluate the effectiveness of CGP, extensive experiments are conducted across 6 GNN models on 14 graph datasets. We demonstrate that our method reliably replicates the performance levels of LTH-based methods while achieving significantly faster inference (e.g., up to $250$\textit{×} faster inference on the Photo dataset). In particular, on high-heterophily benchmark datasets, our method consistently outperforms SOTA hetero-GNN methods, and achieves an improvement of up to $120\%$ over baseline GNN models (GCN~\cite{gcn}, SGC~\cite{sgcn}, and APPNP~\cite{appnp}) while delivering up to nearly $50$\textit{×} faster inference. The main contributions of the present work are as follows:
\begin{enumerate}
  \item We propose a graph gradual pruning framework, namely CGP, to reduce the training and inference computing costs of GNN models while preserving their accuracy.
  \item We comprehensively sparsify the elements of GNNs, including graph structures, the node feature dimension, and model parameters, to significantly improve the efficiency of GNN models.
  \item Experimental results on various GNN models and datasets consistently validate the effectiveness and efficiency of our proposed CGP.
\end{enumerate}

\section{Related Work}
\label{sec:related work}

\textbf{Graph Neural Networks.} Graph Neural Networks (GNNs), which generalize the convolutional operation to graph data, have achieved great success in many real-world applications. The basic idea behind such GNN models as Graph Convolutional Network (GCN)~\cite{gcn}, GraphSAGE~\cite{graphsage},  Graph Attention Network (GAT)~\cite{gat}, and Graph Isomorphism Network (GIN)~\cite{gin}, is to update the embedding of each node with messages from its neighboring nodes. Despite enormous success, the above GNN models are shallow and achieve their best performance with 2-layer models. Such shallow architectures limit the expressive ability, hindering the great potential of GNNs in large-scale real-world graphs~\cite{ogb-dataset}. Therefore, there is an increasing trend towards designing deeper GNN architectures such as ResGCN and GCNII. The ResGCN~\cite{deepgcns,deepgcns-tpami} borrows concepts from CNNs, including residual connections and dilated convolutions, and adapts them to build an extremely deep (56-layer) GCN model. The GCNII~\cite{gcnii} proposes to deepen GNN models to 64 layers with two simple techniques: initial residual and identity mapping. For an in-depth understanding of deeper GNNs, please refer to~\cite{deeper-survey}. Despite the impressive empirical results that deeper GNNs have shown on some large-scale graphs, they stil suffer from high computation costs due to the increased scales of graph data and model parameters used in practice. And the high computation costs limit their deployment in resource-constrained or real-time applications.

\textbf{Graph Sparsification \& Coarsening.} Graph sparsification and coarsening are two ways to reduce the size of a graph in order to speed up graph training. Graph sparsification typically removes the task-irrelevant edges in a graph, while graph coarsening reduces the number of nodes. \textit{1) Graph Sparsification.} Conventional graph sparsification methods~\cite{graph-sparsification-laplacian,graph-sparsification-spectral} are usually unsupervised such that the resulting sparsified graphs aim to preserve graph properties and may not favor downstream tasks. Subsequently, NeuralSparse~\cite{neural-sparsification} proposes a supervised graph sparsification technique, which improves generalization power by learning to remove potentially task-irrelevant edges from input graphs. Similar to NeuralSparse, DropEdge~\cite{dropEdge:} is designed to randomly remove edges in each training epoch, which helps mitigate the over-smoothing issue in the training of deep GNNs. Furthermore, SGCN~\cite{sgcn-admm} attempts to sparsify the adjacency matrix with the help of ADMM optimization~\cite{admm}.  \textit{2) Graph Coarsening.} Current works on graph coarsening mainly focus on preserving different properties, which are related to the spectrum of the original graph and coarse graph~\cite{graph-coarsen-spectrally,graph-coarsen-cut}. In addition, GOREN~\cite{goren} incorporates GNNs into graph coarsening for the first time. Furthermore, Huang \textit{et al.}~\cite{graph-coarsen-scale} attempt to leverage graph coarsening to speedup the training of GNNs in the semi-supervised setting.  For a comprehensive description of graph sparsification and coarsening, please refer to~\cite{summarization-survey,acceleration-survey}. However, all above methods mainly focus on the sparsification of graph structures but ignore model parameters.

\textbf{LTH-based GNN Sparsification.} The lottery ticket hypothesis (LTH)~\cite{lth} states that a sparse subnetwork exists in a dense randomly-initialized network, called the winning ticket, which can be trained to achieve comparable performance to the dense network with the same weight initialization. The LTH has been developed and extended to different fields~\cite{lth-linear,lth-efficient,lth-early-bird,lth-deconstruct,lth-check,lth-ownership,lth-longlife,lth-gans,lth-bert}. Recently, Chen \textit{et al.}~\cite{ugs} extend LTH to GNNs and propose the Graph Lottery Ticket (GLT), which refers to co-sparsifying input graphs and model parameters. Then, You \textit{et al.}~\cite{gebt} demonstrate that the Early-Bird Tickets Hypothesis~\cite{lth-early-bird} still holds for GNNs, and further develop efficient and effective detectors to automatically identify Early-Bird Tickets of GNNs. However, the above two GNN sparsification methods cannot be applied in the inductive learning setting due to the transductive nature of masks. Therefore, Sui \textit{et al.}~\cite{icpg} propose the ICPG framework to endow GLT with the inductive pruning capacity by adopting an AutoMasker strategy. However, the LTH-based methods necessitate fully and iteratively (\textit{up to 20×}) training the dense models, which causes increased computation costs for additional rounds of GNN training. Therefore, to boost both training and inference efficiency, we propose a novel pruning framework to dynamically prune the network parameters in the training stage, which greatly reduces the computational cost and can achieve comparable or even better performance compared to the LTH-based methods.

\section{Methodology}
\label{sec:method}

\subsection{Preliminaries}
\label{sec:preliminary}

\textbf{Notations.}  A graph $\mathcal{G}$ can be represented by an adjacency matrix $\boldsymbol{A} \in \{0, 1\}^{ n \times n}$ and a node feature matrix $\boldsymbol{X} \in \mathbb{R}^{ n \times d}$, where $n$ is the number of nodes,  $d$ is the dimension of node features, and  $\boldsymbol{A}[i, j]=1$ if there exits an edge between node $v_{i}$ and node $v_{j}$, otherwise, $\boldsymbol{A}[i, j]=0$. 

\textbf{Graph Neural Networks.} The basic idea behind GNN models is to update the embedding of each node with messages from its neighboring nodes. Taking GCN~\cite{gcn} as an example, a two-layer GCN model~\cite{gcn} can be formulated as
\begin{equation}
\begin{aligned} 
\boldsymbol{Z} &=f(\boldsymbol{A}, \boldsymbol{X})\\&  =\operatorname{softmax}\left(\hat{\boldsymbol{A}} \operatorname{ReLU}(\hat{\boldsymbol{A}} \boldsymbol{X} \boldsymbol{W}^{(0)}) \boldsymbol{W}^{(1)}\right),
\label{eq:gcn}
\end{aligned}
\end{equation}
where  $\hat{\boldsymbol{A}}=\hat{\boldsymbol{D}}^{-\frac{1}{2}}\left(\boldsymbol{A}+\boldsymbol{I}_{n}\right) \hat{\boldsymbol{D}}^{-\frac{1}{2}}$ is the adjacency matrix normalized by the degree matrix $\hat{\boldsymbol{D}}$ of $\boldsymbol{A}+\boldsymbol{I}_{n}$, $\boldsymbol{Z}$ is matrix of the GCN's predictions,  $\boldsymbol{W} = (\boldsymbol{W}^{(0)}, \boldsymbol{W}^{(1)})$ is the weights of the two-layer GCN model, and $\operatorname{ReLU}$ is an activation function.

\begin{figure*}[!t] 
\setlength{\abovecaptionskip}{-0.1cm}   
\begin{center}
\includegraphics[width=0.7\linewidth]{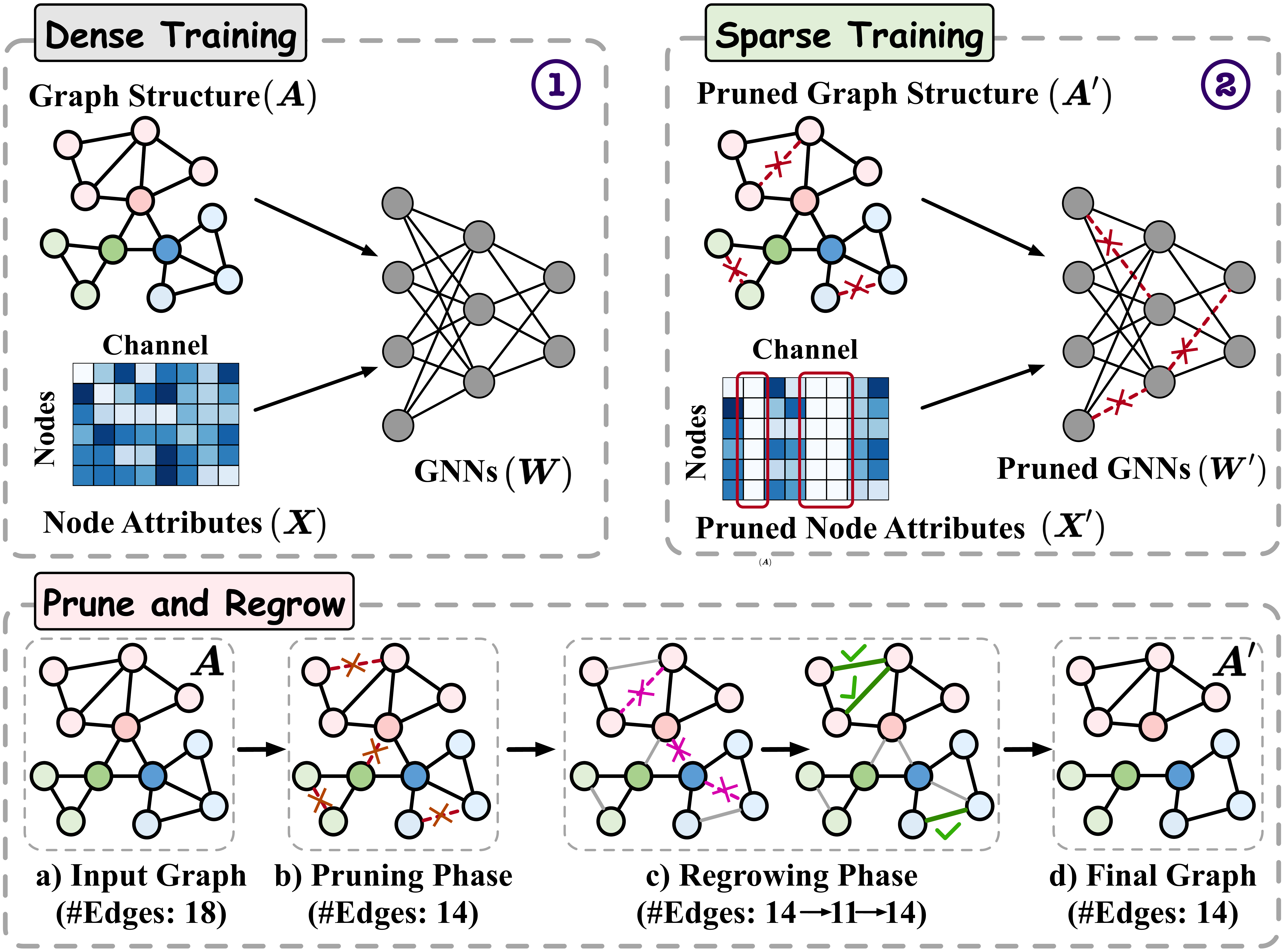}
\end{center}
\caption{An illustration of the proposed graph pruning architecture. In sparse training (\textcolor{blue}{\large \ding{173}}), we prune the unnecessary or unimportant edges, weights, and node feature channels during the training process. \textbf{Bottom.} We take pruning the graph structure ($\boldsymbol{A}$) as an example to illustrate a pruning and regrowth process. The red/pink dashed lines indicate the pruned edges, and the green lines refer to the regrowth edges. Note that the regrowing process does not change the number of edges.}
\label{fig:model}
\end{figure*}

\subsection{Comprehensive Graph Gradual Pruning}

The proposed Comprehensive Graph Gradual Pruning (CGP) architecture is given in Figure~\ref{fig:model}. The CGP can obtain well-performing sparse graphs ($\boldsymbol{A}^{\prime}$ and $\boldsymbol{X}^{\prime}$) and GNN model weights ($\boldsymbol{W}^{\prime}$) in one single training process. In this subsection,  we first introduce the comprehensive sparsification technique which co-sparsifies all three elements found in GNN methods. Next, we describe the gradual magnitude pruning scheme, which performs the sparsification before convergence in the training stage. Finally, a regrowth scheme is given, which helps to recover the pruned critical connections in order to refine the pruning process.

\textbf{Comprehensive Graph Sparsification.} According to Eq.~(\ref{eq:gcn}), there are three elements, $\boldsymbol{A}$,  $\boldsymbol{W}$, and $\boldsymbol{X}$, which can impact the training and inference cost. To this end, our method  simultaneously reduces the edges ($\boldsymbol{A}$) and the feature dimension ($\boldsymbol{X}$) in the graph $\mathcal{G}$ and the model weights ($\boldsymbol{W}$) in GNNs, whereas previous LTH-based methods only focus on co-sparsifying $\boldsymbol{A}$ and $\boldsymbol{W}$.

\textbf{1) We begin with the model weight pruning ($\boldsymbol{W}$).}
Specifically, during the model initialization stage, we create a \textbf{non-differentiable binary mask} $\boldsymbol{m}_{w}$ which is of the same size as the model weights, $\boldsymbol{W}$. The elements in the mask are initially set to unity. At regular intervals (described in the next section), the mask matrix is updated by our pruning strategy that sets the parameters below the threshold to zero, and the weight matrix is multiplied with the updated mask to determine which of the weights participate in the next forward execution of the graph. This procedure can be described as
\begin{equation}
\begin{aligned}
&\mathrm{idx}=\operatorname{TopK}(-|\boldsymbol{W}|,\lceil p_{w} \|\boldsymbol{W} \|_{0} \rceil); \\& \boldsymbol{m}_{w}^{\prime} = \operatorname{Zero}(\boldsymbol{m}_{w}, \mathrm{idx});
 \\& \boldsymbol{W}^{\prime} = \boldsymbol{m}_{w}^{\prime} \odot \boldsymbol{W},  
\label{eq:prune-w}
\end{aligned}
\end{equation}
where $\operatorname{TopK}$ is the function that returns the indices of the top $\lceil p_{w} \|\boldsymbol{W} \|_{0}\rceil$ values in $|\boldsymbol{W}|$, $\operatorname{Zero}$ is the function that sets the values in $\boldsymbol{m}_{w}$ with indices $\mathrm{idx}$ to $0$,  $\boldsymbol{W}^{\prime}$ is the pruned weight matrix,  $p_{w}$ is the sparsity of model weights (i.e., retaining $1-p_{w}$ proportion of weights for the next iteration), $\|\boldsymbol{W} \|_{0}$ is the number of model weights, $\lceil \cdot \rceil$ is the rounding up operator, and $\odot$ is the element-wise product. Note that we adopt the global pruning by default, which refers to pruning the weights of different layers together. 

\textbf{2) Then, we introduce the graph structure pruning ($\boldsymbol{A}$).}
Different from the model weights, which are updated at every optimizer update step, the adjacency matrix is usually fixed along with the training progress. Therefore, we introduce an adjacency attention mask, $\boldsymbol{m}_{a} \in \mathbb{R}^{ \|\boldsymbol{A} \|_{0} }$, a sparse and \textbf{differentiable soft-mask}, where $\|\boldsymbol{A} \|_{0}$ is the number of the edges that we consider for pruning.  After every optimizer update step, the soft-mask, $\boldsymbol{m}_{a}$, is sent to the GNN model, serving as the ``edge-weight'', which is similar to the attention values in GAT~\cite{gat}. At regular intervals, usually in sync with the model weight pruning interval, these masks are updated by setting all parameters that are lower than the threshold to zero, that is
\begin{equation}
\begin{aligned} &
\mathrm{idx}=\operatorname{TopK}(-| \boldsymbol{m}_{a}|,\lceil p_{a} \|\boldsymbol{A} \|_{0} \rceil); \\& \boldsymbol{m}_{a}^{\prime} = \operatorname{Zero}(\boldsymbol{m}_{a},\mathrm{idx}); \\& \boldsymbol{A}^{\prime}_{\text{nonzero}}= \boldsymbol{m}_{a}^{\prime} \odot \boldsymbol{A}_{\text{nonzero}} , 
\label{eq:prune-a}
\end{aligned}
\end{equation}
where $p_{a}$ is the graph structure sparsity,  $\boldsymbol{A}_{\mathrm{nonzero}}$ denotes the edge index\footnote{Edge index denotes the graph connectivity in COO format, which is commonly used in PyG~\cite{pytorch-geometric}.}, $\boldsymbol{A}^{\prime}_{\mathrm{nonzero}}$ is the pruned edge index used in the next training iteration,   and $\cdot_{\mathrm{idx}} $ is an indexing operation.

\textbf{3) Finally, we present the node feature pruning ($\boldsymbol{X}$).}
Besides the graph structure and model weight pruning strategies, we additionally prune the dimension of node features, which is a basic element in all GNNs.  Removing the input features corresponds to removing rows or columns in the weight matrices while sparsifying weights removes the elements of the matrices. Additionally, node features typically have large amounts of redundancy for the downstream tasks (e.g., node classification)~\cite{feature-prune}. Therefore, like graph structure pruning, we introduce a feature attention mask, $\boldsymbol{m}_{x} \in \mathbb{R}^{d}$, a sparse and \textbf{differentiable soft-mask}, where $d$ is the dimension of node features. This soft-mask is applied to the input layer to guide the input node feature pruning. Therefore, the formulation is as follows:
\begin{equation}
\begin{aligned} &
\mathrm{idx}=\operatorname{TopK}(-| \boldsymbol{m}_{x}| ,\lceil p_{x} d \rceil); \\ &
\boldsymbol{m}_{x}^{\prime} = \operatorname{Zero}(\boldsymbol{m}_{x},\mathrm{idx} ); \\& 
 \boldsymbol{X}^{\prime}=\boldsymbol{m}_{x}^{\prime} \odot  \boldsymbol{X},
\label{eq:prune-x}
\end{aligned}
\end{equation}
where $p_{x}$ denotes the sparsity of the feature dimension, and $\boldsymbol{X}^{\prime}$ is the pruned feature matrix used in the next training iteration. Note that in this work we only prune the input features and leave channel pruning (i.e., including embedding pruning) for a future work.

\begin{algorithm}[!t]
\caption{The pseudocode for the proposed CGP.}
\label{alg:CGP}
\begin{algorithmic}[1]
\REQUIRE Graph $\mathcal{G}=\{\boldsymbol{A}, \boldsymbol{X}\}$, GNN $f\left(\mathcal{G}, \boldsymbol{W}\right)$, GNN initialization, $\boldsymbol{W}$, initial masks, $\boldsymbol{m}_{w}$, $\boldsymbol{m}_{a}$, and $ \boldsymbol{m}_{x}$, target sparsity, $p_{w}$, $p_{a}$, and $p_{x}$,  gradual pruning starting point, $t_0$, gradual pruning end point, $t_f$, gradual pruning frequency, $\Delta t$.\\
\FOR{each training step $t$ }  
\STATE {Forward $f\left(\cdot, \boldsymbol{m}_{w} \odot \boldsymbol{W}\right)$ \\ with $\mathcal{G}=\left\{\boldsymbol{m}_{a} \odot \boldsymbol{A}, \boldsymbol{m}_{x} \odot \boldsymbol{X}\right\}$.} 
\STATE {Backpropagate to update $\boldsymbol{W}$, $\boldsymbol{m}_{a}$, and $\boldsymbol{m}_{x}$.}
\IF{$t_o \leq t \leq t_f$  \AND ($t$ $\mathrm{mod}$ $\Delta \mathrm{T}$) == 0} 
 
 \STATE{Gradual pruning $\boldsymbol{W}$, $\boldsymbol{A}$, and $\boldsymbol{X}$ by   Eq.~(\ref{eq:prune-w}),~(\ref{eq:prune-a}), and~(\ref{eq:prune-x}), respectively, with the pruning rate produced by Eq.~(\ref{eq:gradual-pruning}).   }
 \STATE{Regrow $\boldsymbol{W}$, $\boldsymbol{A}$, and $\boldsymbol{X}$ with Eq.~(\ref{eq:regrowth-1}) and Eq.~(\ref{eq:regrowth-2}) }

\ENDIF
\ENDFOR
\end{algorithmic}
\end{algorithm}

\textbf{Gradual Magnitude Pruning.}  The sparsity of candidate elements, introduced in the above section, is often trained with a schedule. According to the previous works~\cite{survey-sparsity,granet}, there mainly exists three different classes of training schedules: train-then-sparsify, before-training-sparsity, and sparsify-during-training. \textbf{1)} The train-then-sparsify schedule performs sparsification after a standard dense training procedure, that is, it runs until convergence in $T$ iterations, followed by a retraining procedure. This usually needs to be performed over many cycles and therefore suffers from an extremely large computation cost. Furthermore, training a dense model to converge may cause over-fitting, which is hard to correct with the subsequent pruning operation~\cite{survey-sparsity}. \textbf{2)} The before-training-sparsity schedule performs sparsification before the main training process, which usually suffers from inadequate performance. \textbf{3)} The sparsify-during-training schedule performs sparsification before the models have been trained to converge ($T$), which can simultaneously exhibit the training/inference efficiency and the comparable performance. 

The LTH-based GNN sparsification methods~\cite{ugs,gebt,icpg} are trained with the train-then-sparsify schedule, resulting in a huge training computation cost. Therefore, we adopt the sparsify-during-training schedule, and for the first time generalize it to GNNs. More specifically, we gradually prune not only the model weighs like in convolutional DNNs, but also the GNNs-specific elements, including the graph structure and node features. The gradual magnitude pruning~\cite{gradual-prune-2,gradual-prune,granet} prunes the elements over $n$ pruning iterations to reach the desired sparsity, as shown in Figure~\ref{fig:compare}. When the sparsifications of all elements are performed every $\Delta t$ steps, then, the pruning rate of each pruning iteration becomes
\begin{equation}
\begin{aligned} 
p_{t}=p_{f}+\left(p_{i}-p_{f}\right)&\left(1-\frac{t-t_{0}}{n \Delta t}\right)^{3}, \\& t \in\left\{t_{0}, t_{0}+\Delta t, \ldots, t_{0}+n \Delta t\right\}, 
\label{eq:gradual-pruning}
\end{aligned}
\end{equation}
where $p_{i}$ is the initial sparsity degree, $p_{f}$ is the target sparsity, $t_{0}$ is the starting epoch of gradual pruning, and $p_{t}$ refers to $p_{w}$, $p_{a}$, or $p_{x}$. The above gradual pruning scheme contributes to a rapid pruning in the initial phase when the redundant connections are abundant and a gradual reduction of connections that are pruned each time as there are fewer and fewer connections remaining. After obtaining the pruning rate, we prune the elements with the smallest magnitude as presented in Eq.~(\ref{eq:prune-w}), (\ref{eq:prune-a}), and (\ref{eq:prune-x}), as this has become the standard method for pruning during the training stage.

\textbf{Regrowth.} Premature pruning may occur during the pruning progress,  especially in the early iterations, causing the loss of important information. To correct the ``mistaken'' pruning, we may incorporate the regrowth schemes, including random regrowth~\cite{growth-set,growth-dsr}, gradient based regrowth~\cite{growth-rigl}, and momentum based regrowth~\cite{growth-snfs}, into the gradual pruning schedule.  In order to ensure that the model remains of approximately the same size, we first remove a proportion of elements before performing regrowth (See part \textbf{c} at the bottom of Figure~\ref{fig:model}). 

Consider regrowing the edges in the graph ($\boldsymbol{A}$) with the regrowth scheme based on gradient~\cite{growth-rigl} as an example. To this end, we first identify the unimportant connections as those with the smallest magnitude (edge weights $\boldsymbol{m}_{a}$), since small magnitude indicates that either the connection’s gradient is small or a large number of oscillations affected the gradient direction. Therefore, these edges have a small contribution to the decrease in training loss and can be removed.  Assuming the regrowth ratio is $r$,  we first remove the $r$-th proportion of unimportant edges (see the left of part \textbf{c} in Figure~\ref{fig:model}) with the smallest magnitude as
\begin{equation}
\begin{aligned} &
\mathrm{idx}=\operatorname{TopK}(-| \boldsymbol{m}_{a} |, r); \\&
 \boldsymbol{A}^{\prime}_{\text{nonzero}}= \operatorname{Zero}(\boldsymbol{A}_{\text{nonzero}}, \mathrm{idx}),
\label{eq:regrowth-1}
\end{aligned}
\end{equation}
where $\boldsymbol{A}^{\prime}_{\text{nonzero}}$ is the edge matrix after removing the $r$-th proportion of edges. Immediately after that, we regenerate the $r$-th proportion of new connections (see the right of part \textbf{c} in Figure~\ref{fig:model}) based on the gradient magnitude, given by
\begin{equation}
\begin{aligned} &
\mathrm{idx} = \operatorname{TopK}\left(\left|\mathbf{g}_{i \notin \boldsymbol{A}^{\prime}_{\text{nonzero}} }\right|, r\right); \\&
\boldsymbol{A}_{\text{nonzero}}^{\prime\prime} =\boldsymbol{A}^{\prime}_{\text{nonzero}} + \operatorname{Zero}(\boldsymbol{A}_{\text{nonzero}}, \mathrm{idx}),   
\label{eq:regrowth-2}
\end{aligned}
\end{equation}
where $\left|\mathbf{g}_{i \notin \boldsymbol{A}^{\prime}_{\text{nonzero}} }\right|$ is the gradient/momentum magnitude of the removed edges, and $\boldsymbol{A}_{\text{nonzero}}^{\prime\prime}$ is the edge matrix after the whole regrowth  process. The regrowth of the other two elements, $\boldsymbol{W}$ and $\boldsymbol{X}$, is the same as that of $\boldsymbol{A}$. We perform the above regrowth operation every $\Delta t$ intervals from the beginning of the training to the end of the pruning. Note that the regrowth scheme is much more training efficient according to the analysis in the previous works~\cite{growth-rigl,granet}.

\textbf{Overall Architecture.} We now describe the full structure of comprehensive graph gradual pruning (CGP), consisting of gradual pruning and regrowth (see Algorithm~\ref{alg:CGP}).  Given a graph, $\mathcal{G} = (\boldsymbol{A}, \boldsymbol{X})$, and a GNN model with initialized weights, $\boldsymbol{W}$, the training (Eq.~(\ref{eq:gcn})) can be compactly represented as
\begin{equation}
\boldsymbol{Z}=f((\boldsymbol{A}, \boldsymbol{X}), \boldsymbol{W}).
\end{equation}
During the training stage, at every $\Delta t$ training steps, we apply the pruning once on $\boldsymbol{A}, \boldsymbol{X}$, and $\boldsymbol{W}$ respectively. The pruning rate is calculated by Eq.~(\ref{eq:gradual-pruning}), and the specific pruning operation is presented in Eq.~(\ref{eq:prune-w}), (\ref{eq:prune-a}), and (\ref{eq:prune-x}). After the pruning operation, we also apply the regrowth operation to ``correct'' the pruning error by Eq.~(\ref{eq:regrowth-1}) and~(\ref{eq:regrowth-2}) before the next round of training.

\textbf{Complexity Analysis.}  
We consider GCN~\cite{gcn} as an example. The time and memory complexities of basic GCN at the inference stage are $\mathcal{O}(L \| \boldsymbol{A} \|_{0} d+$ $\left.L n d^{2}\right)$ and $\mathcal{O}\left(L n d+ L d^{2}\right)$, respectively, where $L, n, \| \boldsymbol{A} \|_{0}$, and $d$ are respectively the total number of GCN layers, nodes, edges, and the dimension of features. For the final sparsities of $ \boldsymbol{W}$, $ \boldsymbol{A}$, and $ \boldsymbol{X}$ as $p_{w}$, $p_{a}$, and $p_{x}$, respectively,  then the inference time and memory complexities of GCN resulting from our proposed CGP are $\mathcal{O}(L (1- p_{a}) \| \boldsymbol{A} \|_{0} (1-p_{x}) d+$ $\left. (1-p_{w})L n ((1-p_{x})d)^{2}\right)$ and $\mathcal{O}\left(L n  (1-p_{x}) d+ (1-p_{w}) L ((1-p_{x}) d)^{2}\right)$, respectively.

\textbf{Further Discussion.} To the best of our knowledge, the methods most relevant  to our CGP are the LTH-based GNN sparsification methods, including UGS~\cite{ugs}, GEBT~\cite{gebt}, and ICPG~\cite{icpg}. The original motivation for these methods is to trim down the inference Multiplication-and-ACcumulation operations (MACs) without sacrificing performance. More specifically, LTH-based methods attempt to identify the Graph Lottery Ticket (GLT), consisting of a subnetwork ($\boldsymbol{W}_{\text{sparse}} \subset \boldsymbol{W}$) and a sparse graph ($\boldsymbol{A}_{\text{sparse}} \subset \boldsymbol{A}$), which maintains the performance of dense networks on the full graph. To achieve the above goal, they usually adopt an iterative training cycle (see Figure~\ref{fig:compare} \textit{left}): \textbf{1)} train-dense, \textbf{2)} prune, \textbf{3)} re-train, until the desired sparse rates for the graph structures and model weights are respectively reached.  While the LTH-based methods trim down the inference MACs, they suffer from a huge training cost since they require many pruning and re-training cycles to achieve the desired performance (up to 20x).  Our CGP method, however, can obtain well-performing sparse graphs ($\boldsymbol{A}^{\prime}$ and $\boldsymbol{X}^{\prime}$) and GNN model weights ($\boldsymbol{W}^{\prime}$) with one single training process (see Figure~\ref{fig:compare} \textit{right}). Therefore, the CGP exhibits training and inference efficiency to dense GNNs while maintaining comparable performance of GNN models.

\section{Experiments}
\label{sec:experiment}

To validate the effectiveness of the proposed method, we conducted extensive experiments across 6 GNN architectures on 14 benchmark graph datasets. From the experiment results, we obtained \textbf{11 observations} in total.

\begin{figure*}[!t] 
\begin{center}
\includegraphics[width=1.0\linewidth]{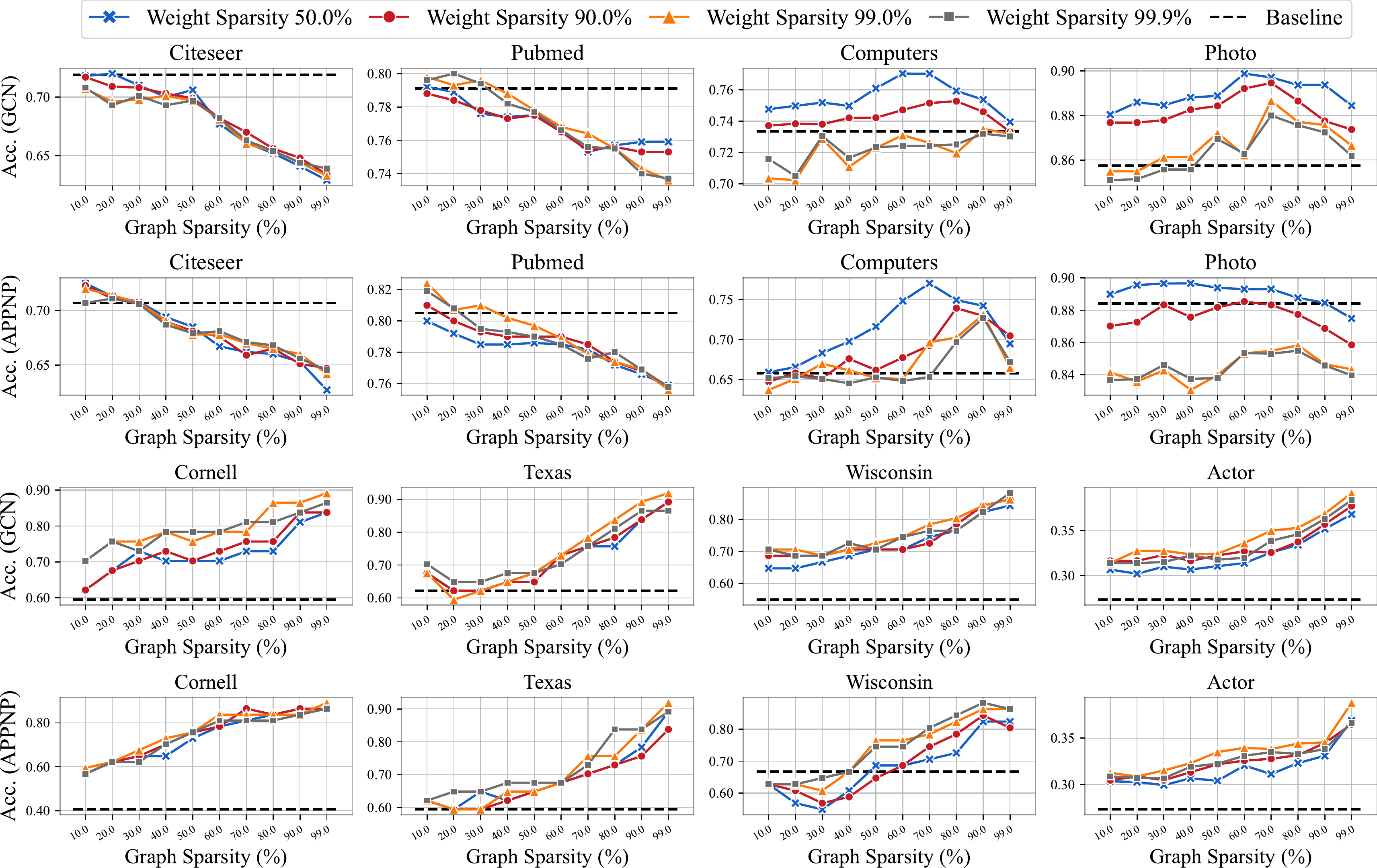}
\end{center}
\caption{Results of co-sparsifying model weights and graph structures, with the proposed \textbf{CGP} compared with backbone models (GCN and APPNP) over 8 graph datasets. \textbf{Top \& left}: Small-scale, high-homophily (low-heterophily), and small node degrees. \textbf{Top \& Right}:  Middle-scale, high-homophily (low-heterophily), and large node degrees. \textbf{Bottom}:  Small-scale, high-heterophily.  More results (GAT and SGC on more datasets) are referred to Appendix~\ref{sec:apd-vs-base}.}
\label{fig:main-wa}
\end{figure*}

\begin{figure*}[!t] 
\begin{center}
\includegraphics[width=1.0\linewidth]{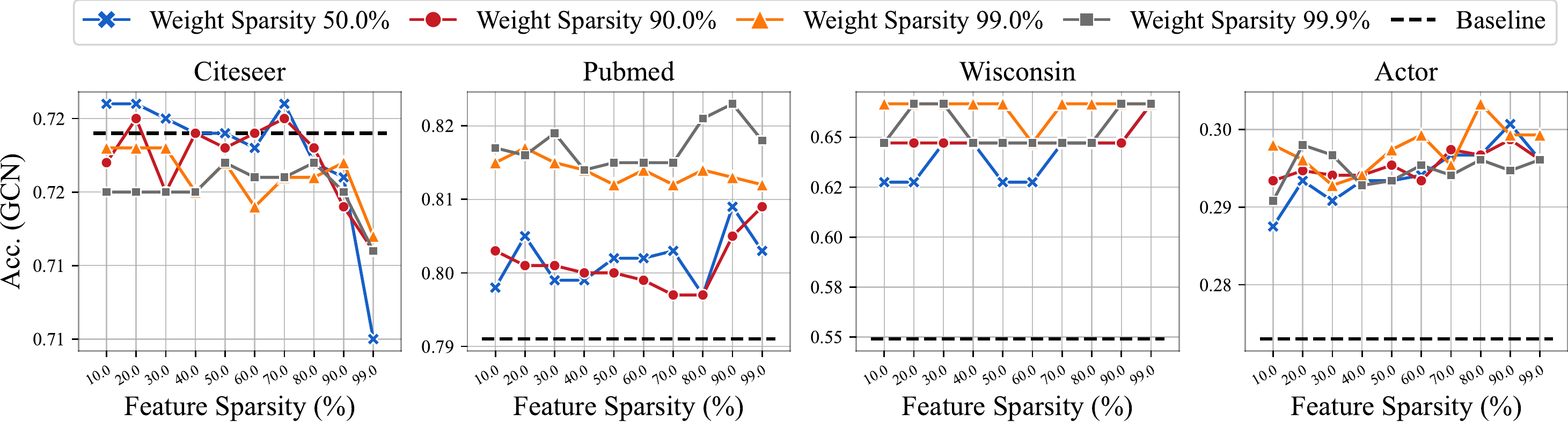}
\end{center}
\caption{Results of co-sparsifying model weights and node features (GCN). }
\label{fig:main-wf}
\end{figure*}

\subsection{Experimental Settings}

\textbf{Datasets.} We used a total of 14 real-world datasets, including 7 small-scale graphs (Cora, CiteSeer, PubMed, Cornell, Texas, Wisconsin, and Actor), 4 medium-scale graphs (CS, Physics, Photo, and Computers), and 3 large-scale graphs (ogbn-arxiv, ogbn-proteins, and ogbn-products)  from Open Graph Benchmark~\cite{ogb-dataset}, which involve diverse domains (citation, co-authorship, co-purchase, web page, and biology). We used their original settings (e.g., splitting) wherever possible.  The detailed statistics of the above datasets are described in Table~\ref{tab:dataset} of Appendix~\ref{sec:apd-dataset}. More detailed descriptions and references can also be found in Appendix~\ref{sec:apd-dataset}.

\textbf{Baseline Models.} \textbf{1) GCN.} \textbf{2) GAT.} They are two representative \textit{shallow} graph neural models~\cite{gcn,gat}, over which we evaluate the proposed method on small- and medium-scale datasets. \textbf{3) SGC.} \textbf{4) APPNP.} Two \textit{shallow-but-deep-propagation} models~\cite{sgcn,appnp}. \textbf{5) ResGCN.}  \textbf{6) GCNII.}  They are two \textit{deeper} graph neural models~\cite{deepgcns,gcnii}, over which we conducted experiments with some large-scale datasets like ogbn-arxiv, ogbn-proteins, and ogbn-products. Please refer to Appendix~\ref{sec:apd-model} for the detailed introductions.

\textbf{SOTA Sparsification Methods for GNNs.} \textbf{7) UGS.} This method~\cite{ugs} extends LTH to GNNs and proposes the Graph Lottery Ticket (GLT). \textbf{8) GEBT.} This method~\cite{gebt} extends the Early-Bird Tickets Hypothesis~\cite{lth-early-bird} to GNNs. \textbf{9) ICPG.}  This method~\cite{icpg} endows the GLT with the inductive learning.

\textbf{SOTA Non-homophilous Methods.} To evaluate the effectiveness of our proposed method on heterophilous graphs, we further compare it with several benchmarking non-homophilous methods. \textbf{10)  MLP.}  The method~\cite{deep-learning} that only uses node features. \textbf{11)  LINK.} The method (logistic regression on the adjacency matrix)~\cite{logistic}) that only uses the graph topology. \textbf{12) H2GCN.} \textbf{13) MixHop.} \textbf{14) GPRGNN.} \textbf{15) FAGCN.}  Recent methods~\cite{h2gcn,mixhop,gprgnn,fagcn} designed for non-homophilous graphs. See Appendix~\ref{sec:apd-model} for the detailed descriptions of the above  9 SOTA methods. 

\textbf{Implementation Details.} We evaluated the effectiveness of the proposed model (improved training and inference efficiency) in terms of the node classification accuracy, inference FLOPs, and total training FLOPs. For shallow and shallow-but-deep-propagation models, we followed~\cite{gcn} to train all the chosen two-layer GNN models on small- and medium-scale graph datasets. For deep models, we followed~\cite{deepgcns} to train ResGCN and GCNII on OGB graphs. Experimental details including all the training hyper-parameters are described in Appendix~\ref{sec:apd-exp-setting}.

\begin{figure}[!t] 
\begin{center}
\includegraphics[width=1.0\linewidth]{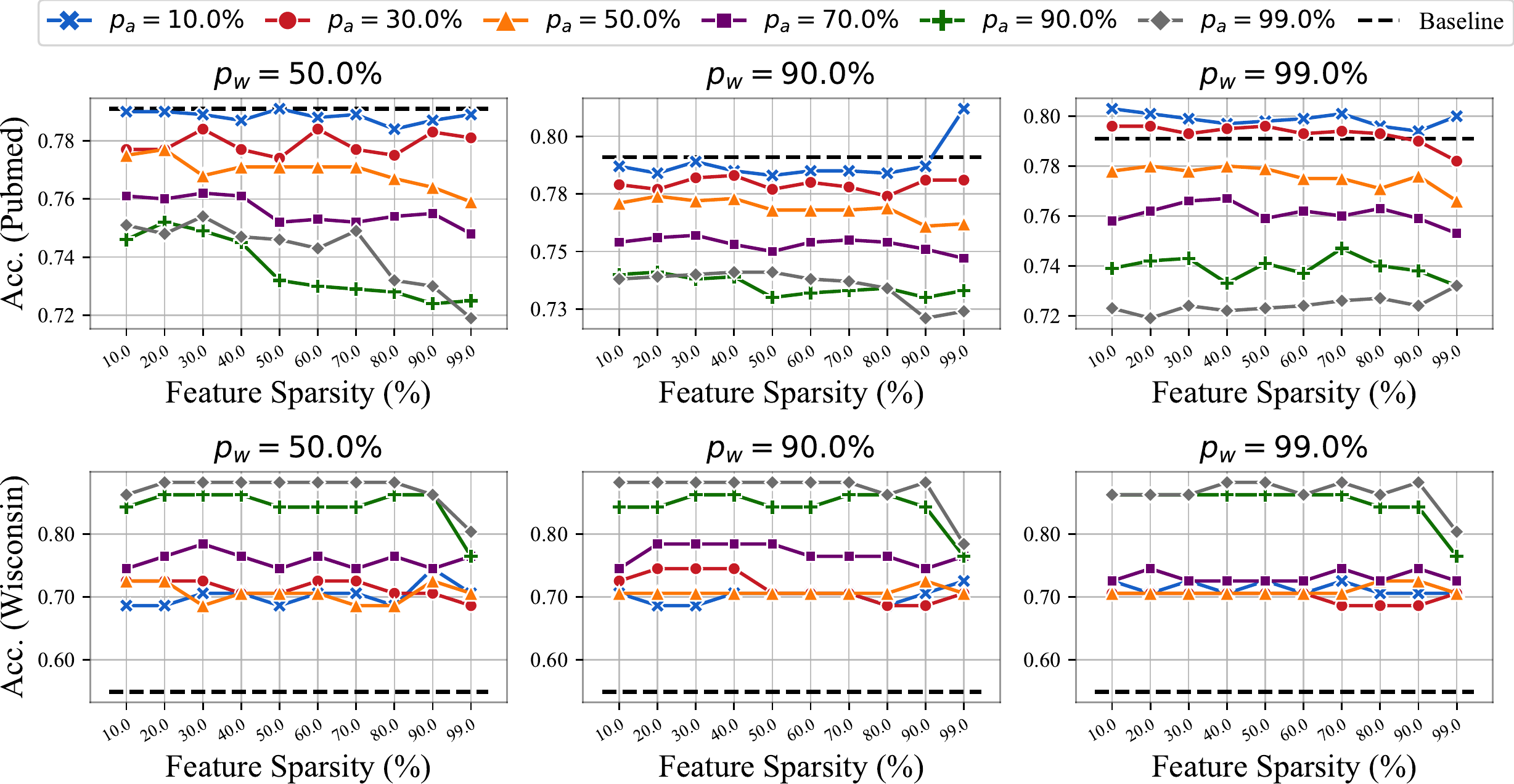}
\end{center}
\caption{Results of co-sparsifying model weights, graph structures, and node features (GCN).  More results are given in Appendix~\ref{sec:apd-vs-base}.}
\label{fig:main-waf}
\end{figure}

\subsection{Shallow GNNs on Small and Medium-scale Datasets}

\subsubsection{Proposed CGP vs. Baseline Models}

To evaluate the benefits of our proposed CGP, we first compared the performance of CGP over four commonly used GNN models on 10 benchmark graph datasets from three perspectives: \textbf{1)} Co-sparsifying model weights and graph structures ($\boldsymbol{W} \& \boldsymbol{A}$). \textbf{2)}  Co-sparsifying model weights and node features ($\boldsymbol{W} \& \boldsymbol{X}$). \textbf{3)}  Co-sparsifying model weights, graph structure, and node features ($\boldsymbol{W} \& \boldsymbol{A} \& \boldsymbol{X}$). The results of the three perspectives are shown in  Figures~\ref{fig:main-wa},~\ref{fig:main-wf}, and~\ref{fig:main-waf}, respectively. Note that when sparsifying $\boldsymbol{W} \& \boldsymbol{A}$ / $\boldsymbol{W} \& \boldsymbol{X}$, we keep $\boldsymbol{X}$/$\boldsymbol{A}$ complete.  Due to space limitation, we only present parts of the results in the main text, and the rest can be found in Appendix~\ref{sec:apd-vs-base}. From the results, we obtain the following \textbf{observations}.

\textbf{Obs.1. The extreme sparsity of model weights can still maintain satisfactory performance.} As shown in Figure~\ref{fig:main-wa}, the performance of $99.9 \%$ weight sparsity (i.e., only 0.1\% of model weights remain at the inference stage) is consistently comparable to that of low-level weight sparsity (e.g., $99.0 \%$, $90.0 \%$, and $50.0 \%$) over different graph sparsities on most models and datasets. Furthermore, on several small-scale datasets (e.g., Texas and Wisconsin), the extreme sparsity of model weights can maintain higher accuracies compared with other sparsities. And we can also obtain the same observation in Figures~\ref{fig:main-wf} and~\ref{fig:main-waf}.

\textbf{Obs.2. Denser graphs are more resilient to graph sparsification.} As shown in Figure~\ref{fig:main-wa}, the performance on the Computers and Photo datasets does not decrease, as the sparsity becomes larger in most cases. According to our analysis, this can be attributed to the densities of graph datasets. Specifically, Computers (Avg. Node Degree: 35.76) and Photo (Avg. Node Degree: 31.13) are significantly denser than Citeseer (Avg. Node Degree: 2.78) and Pubmed (Avg. Node Degree: 4.50).  Therefore, the neighborhood information of each node may be redundant for the prediction, and thus the edges which propagate the information can be cut off by a large amount. Furthermore, we discovered that the model usually achieves the best performance when the sparsity of the graph structure is about 50\%, which indicates that about half of the edges are redundant for the downstream task (i.e., node classification).

\textbf{Obs.3. CGP achieves a great improvement on heterophily graphs.} In a heterophily (or low homophily) graph (e.g., Cornell), the connected nodes may have different class labels and dissimilar features, with a high probability. Therefore, most edges can be seen as ``noisy'' edges for the node classification task and should be cut off. As shown in Figure~\ref{fig:main-wa}, the model performance becomes better as the graph sparsity becomes higher (i.e., cutting-off more edges), which supports our above analysis. Note that different from MLP, which does not utilize the edges from beginning to end, our method gradually reduces the number of edges, and we do not cut off all edges. The above two conditions enable our method to take full advantage of the ``truth'' edges. We also present a detailed comparison of MLP and our CGP in Section~\ref{sec:vs-sota-hetero}.

\textbf{Obs.4. The node feature channels in most graphs are redundant.} The dimension of node feature vectors in graphs can be very high, e.g., each node in the Citeseer graph has 3,703 features. As shown in Figure~\ref{fig:main-wf}, the model usually achieves the best performance when the feature sparsity is about 70\% or 80\%, which indicates that the model does not require too many features for each node when performing inference. Therefore, by aggressively trimming down the node features, performance can still be maintained or even be improved. In addition, sparsifying node features could also significantly improve the memory efficiency by reducing the cache miss rate and peak memory usage under the Gather-ApplyEdge-Scatter platforms (e.g., Pytorch-Geometric~\cite{pytorch-geometric})~\cite{feature-efficiency}.

\textbf{Obs.5. Graph structure is more sensitive to pruning.} As shown in Figure~\ref{fig:main-waf}, when we co-sparsify $\boldsymbol{W} \& \boldsymbol{A} \& \boldsymbol{X} $, the graph structure, $\boldsymbol{A}$, always plays a more important role in the model performance. Specifically, the gap between the performance of different graph sparsities is larger than that of different feature sparsities and weight sparsities on Pubmed and Wisconsin. The former is a homophilic graph, and the latter is a heterophilic one. The same observation can also be obtained when we co-sparsify $\boldsymbol{W} \& \boldsymbol{A}$, as shown in Figure~\ref{fig:main-wa}. More experiments are presented on other datasets in Appendix~\ref{sec:apd-vs-base}, which can also support our above discussion.

\subsubsection{Proposed CGP vs. SOTA Sparsification Methods for GNNs}

\begin{figure*}[!t] 
\begin{center}
\includegraphics[width=1.0\linewidth]{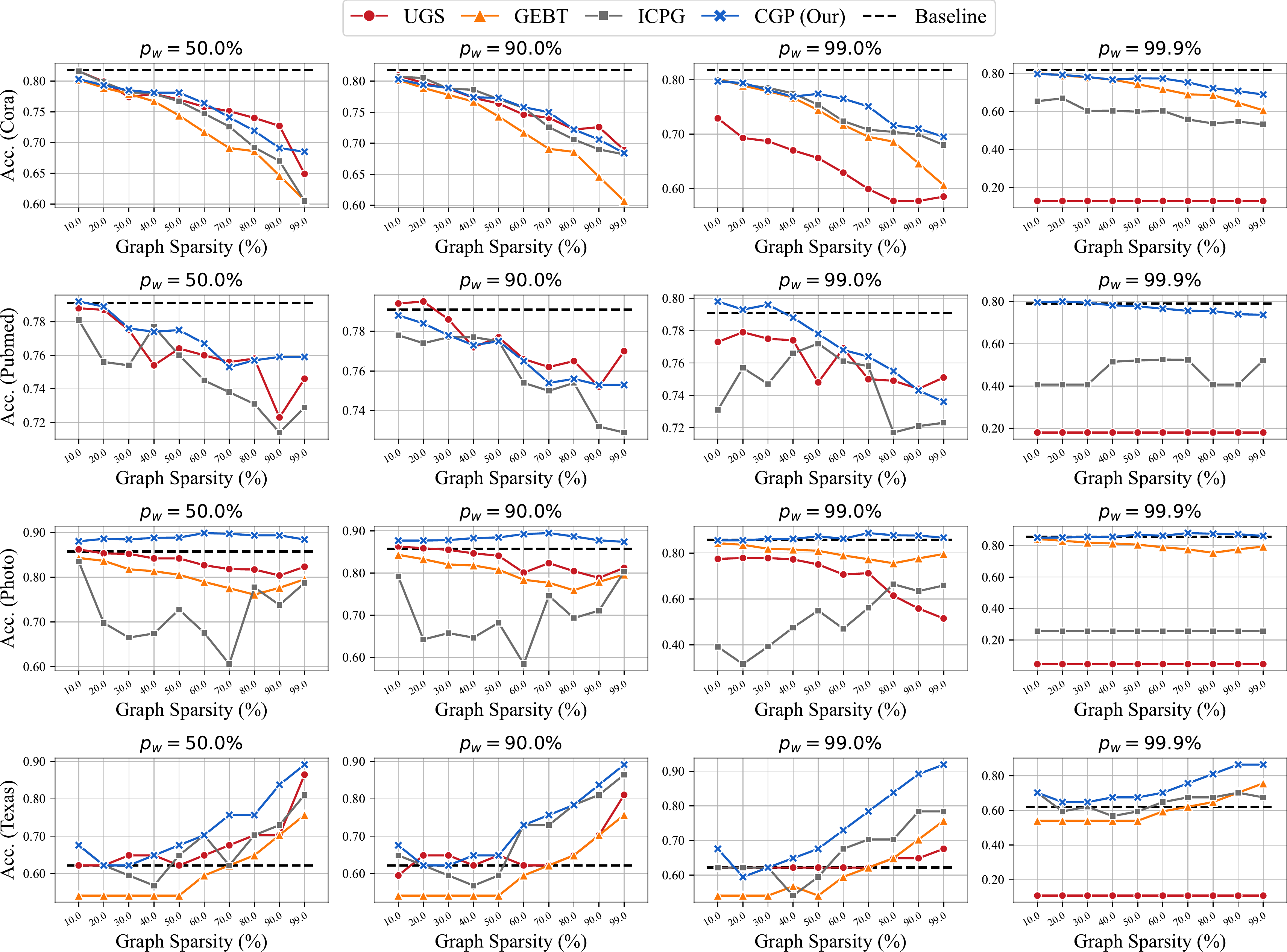}
\end{center}
\caption{A comparison of the proposed CGP framework with SOTA sparsification methods for GNN models. More results are demonstrated in Appendix~\ref{sec:apd-exp-sota-gnn}. }
\label{fig:main-vs-sota}
\end{figure*}

\begin{figure*}[!t] 
\begin{center}
\includegraphics[width=1.0\linewidth]{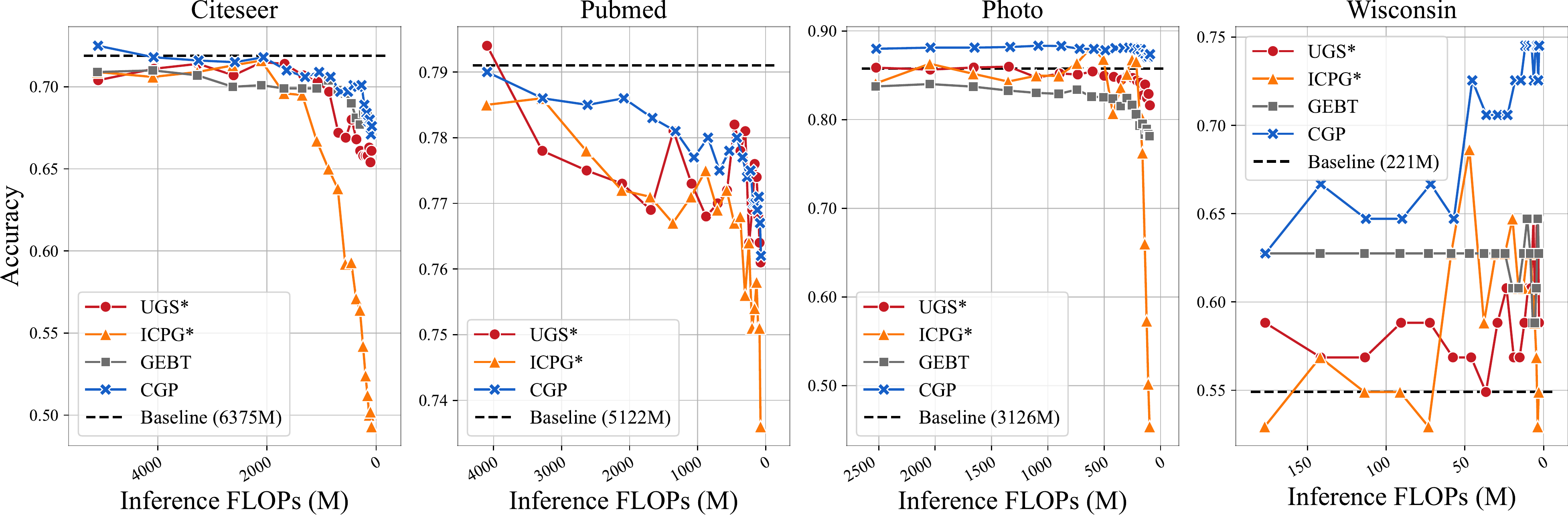}
\end{center}
\caption{A comparison of the proposed CGP framework with SOTA sparsification methods for GNN models in the context of FLOPs. More results are given in Appendix~\ref{sec:apd-exp-sota-gnn}. }
\label{fig:main-flops}
\end{figure*}

\begin{figure}[!t] 
\begin{center}
\includegraphics[width=1.0\linewidth]{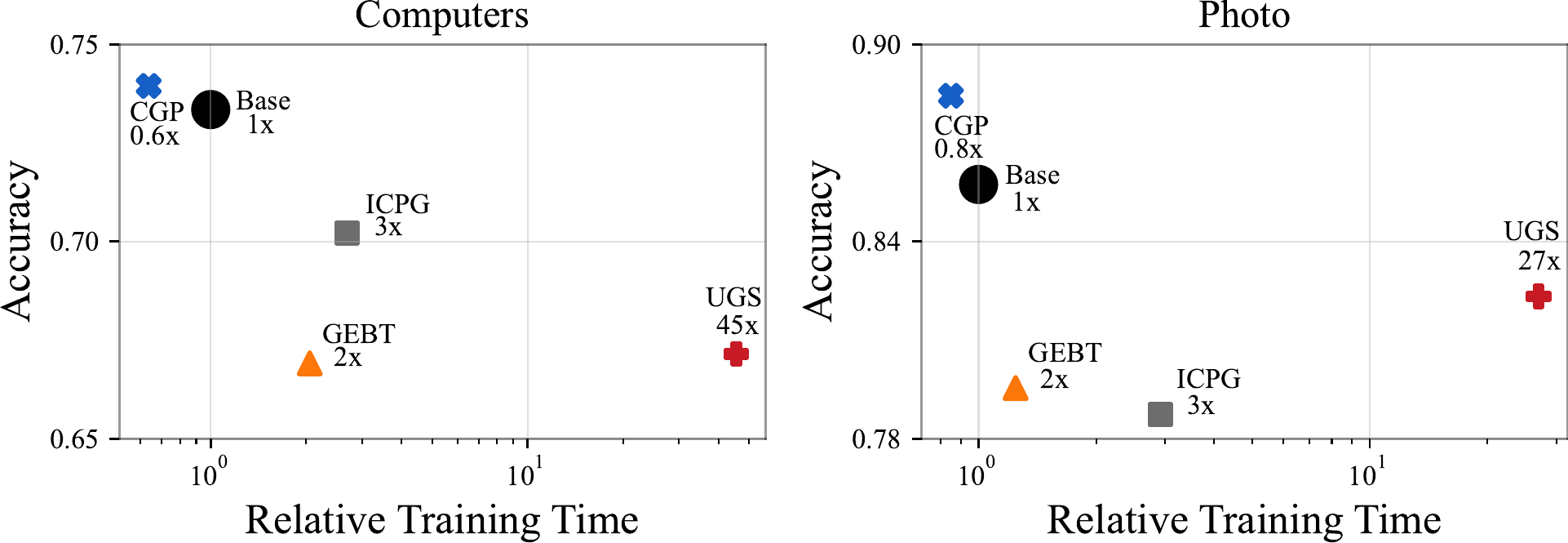}
\end{center}
\caption{Performance over training time on two graph datasets (Computers and Photo). Our method is faster than LTH-based methods while achieving competitive performance. }
\label{fig:main-train-time}
\end{figure}

To further explore the effectiveness of the proposed CGP, we compared CGP with several SOTA LTH-based GNN sparsification pipelines, including UGS~\cite{ugs}, GEBT~\cite{gebt}, and ICPG~\cite{icpg}. For a fair comparison, we adopted two different settings, one-shot pruning and iterative pruning, for UGS and ICPG methods. Note that our CGP and GEBT methods always adopt one-shot pruning, which is much more effective than iterative pruning during the training process. The comparison results are shown in Figures~\ref{fig:main-vs-sota},~\ref{fig:main-flops}, and~\ref{fig:main-train-time}. Due to space limitation, we only present partial results in the main text, and the rest can be found in Appendix~\ref{sec:apd-exp-sota-gnn}. According to the results, we summarize the \textbf{observations} below.

\textbf{Obs.6. CGP consistently outperforms SOTA LTH-based methods.} As shown in Figure~\ref{fig:main-vs-sota}, all methods adopt one-shot pruning, using only one pruning step to reach the final desired sparsity. We can observe that CGP consistently surpasses SOTA LTH-based methods by substantial performance margins across all datasets and GNNs, which validates the effectiveness of our proposed framework. In Figure~\ref{fig:main-flops}, UGS and ICPG adopt iterative pruning, which requires a comprehensive and iterative (up to 20\textit{×}) training of dense models to reach the desired sparsity, while our CGP and GEBT methods continue to adopt one-shot pruning. Therefore, our CGP method requires the smallest number of training FLOPs while achieving a comparable performance across all datasets.

\textbf{Obs.7. CGP is more amenable to the extreme sparsity than SOTA LTH-based methods.} In Figures~\ref{fig:main-vs-sota} and~\ref{fig:main-flops}, the performance of our method does not degrade significantly with the increase of sparsity rates (the decrease of the inference FLOPs) compared with the SOTA methods, which suggests that CGP is more amenable to the extreme sparsity.  Especially when the weight sparsity reaches 99.9\% (only 0.1\% edges of graph remain at the inference stage),  UGS and ICPG completely fail to handle this case, limiting their applications in the resource-constrained situations.

\textbf{Obs.8. CGP substantially saves inference MACs.} As shown in Figure~\ref{fig:main-flops}, when the inference FLOPs (1 FLOPs $\approx$ 2 MACs) reaches about only $100$ M, up to $99\%$ MACs saving of the baseline models, CGP can still provide satisfactory performance and maintain higher accuracies on severely sparsified graphs.

\textbf{Obs.9. CGP better boosts training efficiency compared to SOTA LTH-based methods.} Figure~\ref{fig:main-train-time} shows the performance of the GNN sparsification methods over their training time relative to that of base GCN on the Computers and Photo datasets. For a fair comparison, we set all methods with the same sparsification rate and objects ($\boldsymbol{A}$ and $\boldsymbol{W}$). Observe that the training time of UGS is about $45$\textit{×} bigger than that of the baseline model (GCN), and that our method can be trained up to about two orders of magnitude faster than LTH-based methods while maintaining higher accuracies.

\begin{table}[!t]
\renewcommand\arraystretch{1.2} 
\setlength\tabcolsep{1pt} 
\caption{Our CGP vs. SOTA heterophilous GNN methods. \textbf{Bold}: the best performance per benchmark. \underline{Underline}: the second best performance per benchmark.}
\label{tab:heter-result}
\resizebox{0.5\textwidth}{!}{%
\begin{tabular}{@{}lcccc@{}}
\toprule
                      & \textbf{Cornell}         & \textbf{Texas}           & \textbf{Wisconsin}       & \textbf{Actor}           \\ \midrule
GCN~\cite{gcn}                   & 59.46           & 62.16           & 54.90           & 30.30           \\
SGC~\cite{sgcn}                   & 56.76           & 59.46           & 54.90           & 26.84           \\
APPNP~\cite{appnp}                 & 40.54           & 59.46           & 66.67           & 27.30           \\ \midrule
LINK~\cite{logistic}                  & 59.16           & 78.38           & 66.67           & 22.83           \\
MLP~\cite{deep-learning}                   & 78.38           & 78.38           & 74.51           & 35.13           \\
H2GCN~\cite{h2gcn}                 & 78.38           &81.08          & \underline{84.31}           & 35.66           \\
MixHop~\cite{mixhop}                & 56.76           & 64.86           & 66.67           & 27.57           \\
GPRGNN~\cite{gprgnn}                & \underline{83.78}           & 78.38           & \textbf{88.24}          & 36.38           \\
FAGCN~\cite{fagcn}                 & 81.08           & 72.97           & 72.55           & \underline{36.51}            \\ \midrule
\textbf{Our (GCN) vs. GCN}     & \textbf{89.19 ($\uparrow$50.0\%)}  & \underline{89.19 ($\uparrow$43.4\%)}  & \textbf{88.24 ($\uparrow$60.7\%)}  & \textbf{39.21 ($\uparrow$29.4\%)}  \\
\textbf{Our (SGC) vs. SGC}    & 56.76 (0.0\%)   & 67.57 ($\uparrow$13.6\%)  & 58.82 ($\uparrow$7.1\%)   & 27.70 ($\uparrow$3.2 \%)  \\
\textbf{Our (APPNP) vs. APPNP} & \textbf{89.19 ($\uparrow$120.0\%)} & \textbf{91.89($\uparrow$54.5\%)} & \textbf{88.24 ($\uparrow$32.4\%)} & 38.82 ($\uparrow$42.2 \%) \\ \bottomrule
\end{tabular}%
}
\end{table}

\begin{figure*}[!t] 
\begin{center}
\includegraphics[width=1.0\linewidth]{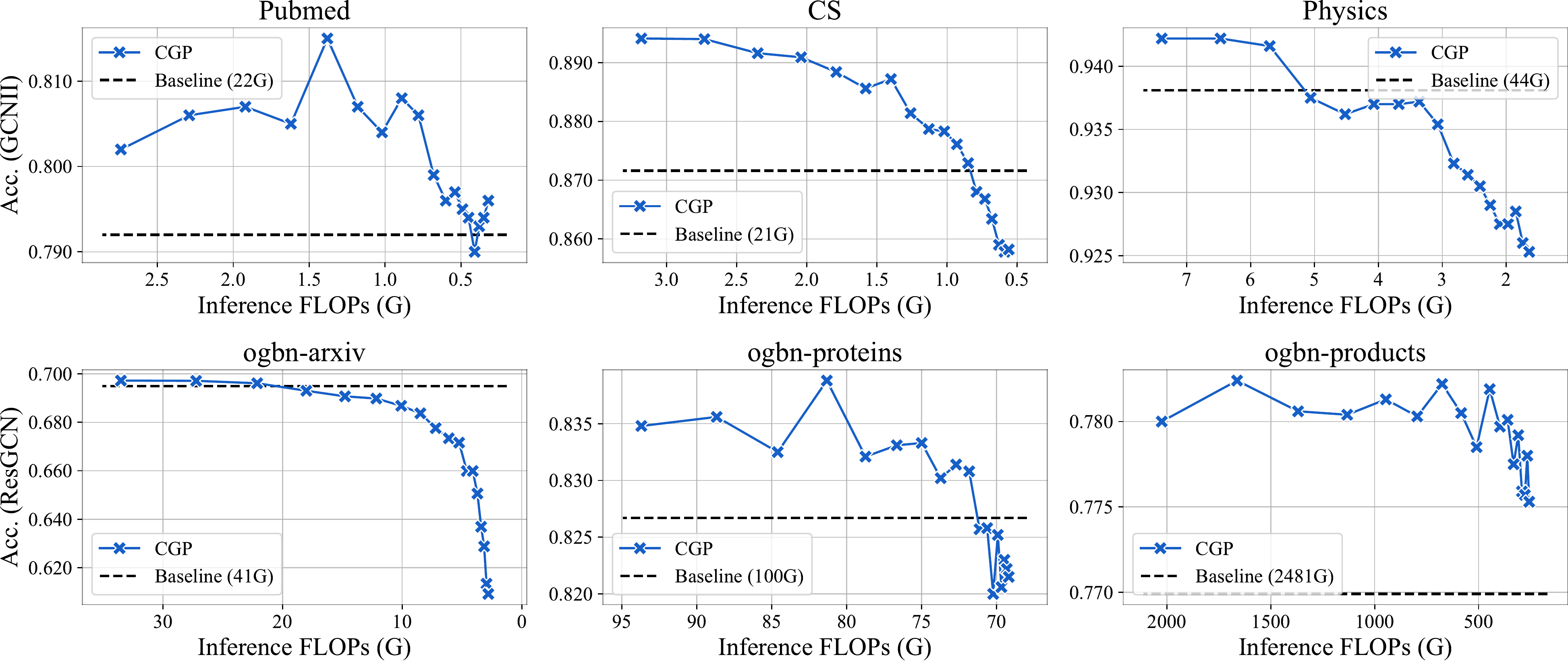}
\end{center}
\caption{Results of sparsifying deeper GNN models.}
\label{fig:main-deep}
\end{figure*}

\subsubsection{Proposed CGP vs. SOTA Hetero-GNN Methods}
\label{sec:vs-sota-hetero}

Furthermore, we compared the CGP with several SOTA GNN methods, which were recently proposed for non-homophilous settings on heterophilic graphs (H2GCN~\cite{h2gcn},  MixHop~\cite{mixhop},  GPRGNN~\cite{gprgnn}, and FAGCN~\cite{fagcn}).  The results are given in Table~\ref{tab:heter-result}. According to the results, we obtain the following \textbf{observation}.

\textbf{Obs.10. CGP consistently performs better than Hetero-GNN methods.} We conclude the results on heterophilic graph datasets in Table~\ref{tab:heter-result}, where $\uparrow$ denotes an improvement over the original baseline models. It can be seen that CGP consistently outperforms all the baselines and SOTA methods in terms of accuracy. Specifically, CGP achieves up to $120\%$ improvements as compared to baselines (GCN~\cite{gcn}, SGC~\cite{sgcn}, and APPNP~\cite{appnp}). Furthermore, compared with MLP, which cuts off all edges, our method also consistently reaches higher accuracies on all heterophilic graph datasets.  One possible explanation is that our method gradually reduces the number of edges, which means that we can take advantage of the useful edges during the training stage. We also applied our method to SOTA Hetero-GNN methods, such as H2GCN~\cite{h2gcn} or FAGCN~\cite{fagcn}, to further improve their efficiency and accuracy. However, this is not the key point of our work, and the results are presented in Appendix~\ref{sec:apd-exp-sota-hetero}.

\begin{figure*}[!t] 
\begin{center}
\includegraphics[width=1.0\linewidth]{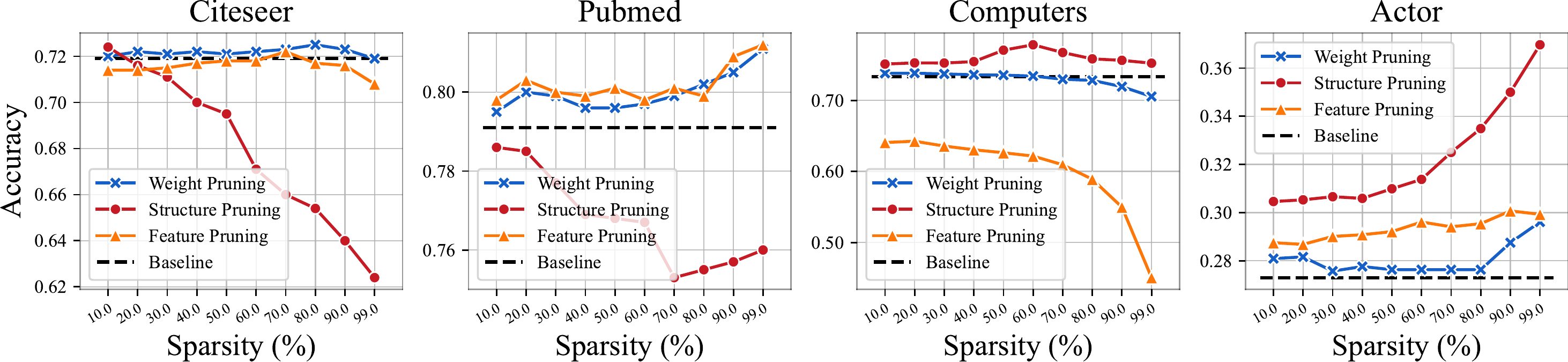}
\end{center}
\caption{Performance of various pruning strategies for the ablation study. For more results, we refer to Appendix~\ref{sec:apd-exp-ablation}.}
\label{fig:main-ablation}
\end{figure*}

\subsection{Scale to Deeper GNNs}

In this section, we further conduct experiments with deeper GNNs (ResGCN~\cite{deepgcns} and GCNII~\cite{gcnii}), which induce a large amount of weights ($\boldsymbol{W}$) on small- and large-scale datasets. Specifically, we tested GCNII (64-layer) on 3 small-and medium-scale datasets (Pubmed, CS, and Physics), and ResGCN (28-layer) on 3 large-scale datasets (ogbn-arxiv, ogbn-products, and ogbn-proteins), which have a large amount of edges ($\boldsymbol{A}$). Our focus was on sparsifying $\boldsymbol{A}$ and $\boldsymbol{W}$. The results are illustrated in Figure~\ref{fig:main-deep} and summarized in the following observation.

\textbf{Obs.11. CGP is scalable for deep GNNs.} Figure~\ref{fig:main-deep} demonstrates that CGP can be scaled up to deep GNNs. \textbf{1)} CGP achieves  matching performance with base GCNII with about $95\%$ and $90\%$ MACs saving on CS and Physics datasets, respectively. Also, CGP almost outperforms base GCNII on Pubmed (up to 98\% MACs saving). \textbf{2)} CGP obtains matching performance with base ResGCN with $50\%$ and $25\%$ MACs saving on ogbn-arxiv and ogbn-proteins, respectively. Furthermore, our CGP consistently outperforms base ResGCN on ogbn-products (up to 80\% MACs saving).\footnote{Our results of ResGCN on OGB datasets (in Figure~\ref{fig:main-deep}) are worse than those of leaderboards, since we cannot perform full batch testing like in leaderboards, which takes about 405 G RAM to do inference on the whole graph. In this paper, we performed mini-batch testing by partitioning the graph, and therefore the resulting accuracy suffered from a decrease.}

\subsection{Ablation Study}

\textbf{Spasifying $\boldsymbol{W}, \boldsymbol{A}, \boldsymbol{X} $ Separately.} 
In this section, we separated the sparsifying operations and explored their roles when applying them to the graphs and the model independently. From Figure~\ref{fig:main-ablation} and Figure~\ref{fig:apd-ablation} in Appendix~\ref{sec:apd-exp-sensitive}, we gained the following interesting observations. \underline{First,} sparsifying the model weights usually helps improve the model performance.  \underline{Second,} in most graphs, the trends of different sparsifying ways are inconsistent, which requires us to find a balance point.

\begin{figure}[!t] 
\begin{center}
\includegraphics[width=1.0\linewidth]{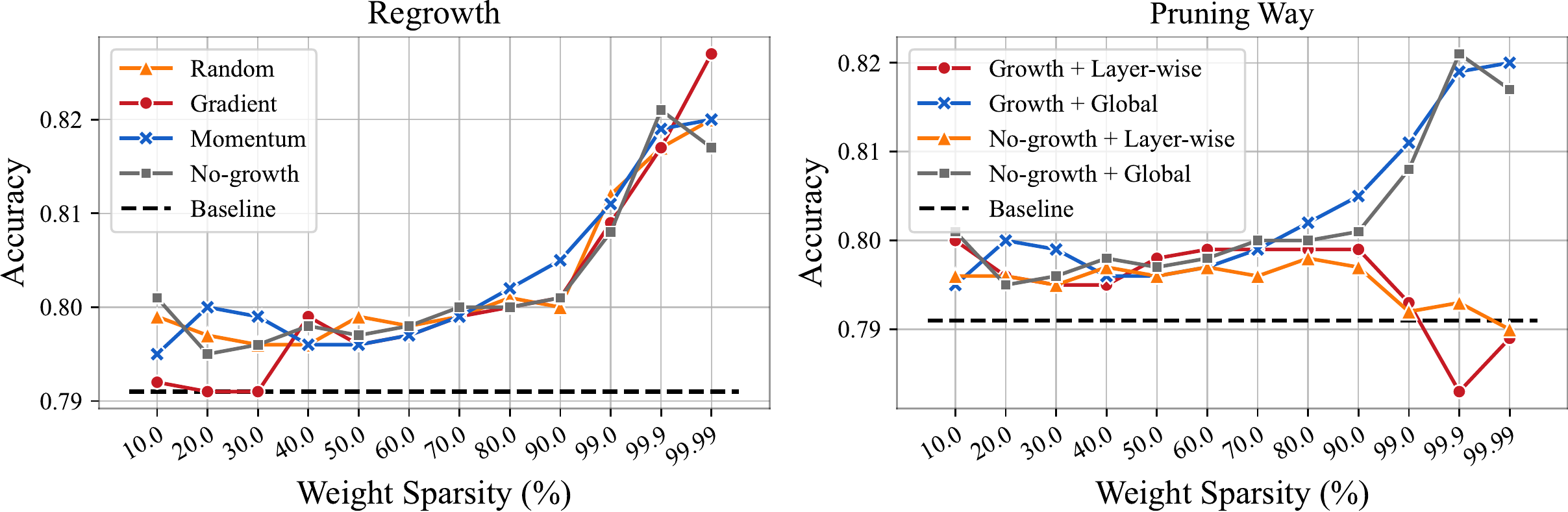}
\end{center}
\caption{Effectiveness of regrowth \& pruning way. More results are given in Appendix~\ref{sec:apd-exp-ablation}. }
\label{fig:main-effectiveness}
\end{figure}

\textbf{Effectiveness of Regrowth \& Pruning Ways.} In this section, we investigated the influence of different regrowth ways, including random growth~\cite{growth-dsr,growth-set}, momentum-based growth~\cite{growth-snfs}, gradient-based growth~\cite{growth-rigl}, and no growth. From Figure~\ref{fig:main-effectiveness} (\textbf{left}), we can observe that all the regrowth mechanisms help boost the performance, especially when the model becomes extremely sparse. Also, the momentum-based regrowth usually performs better than other methods. In addition, we also explored the rules of the pruning way, including layer-wise pruning and global pruning. More specifically, global pruning prunes different layers together and leads to non-uniform sparsity distributions, while layer-wise pruning performs the operation layer by layer, resulting in uniform distributions.  From Figure~\ref{fig:main-effectiveness} (\textbf{right}), we can obviously find that as the model weights become sparser, the global pruning way performs better than the layer-wise pruning. This may be because the global pruning can preserve more connections in important layers.

\begin{figure}[!t] 
\begin{center}
\includegraphics[width=1.0\linewidth]{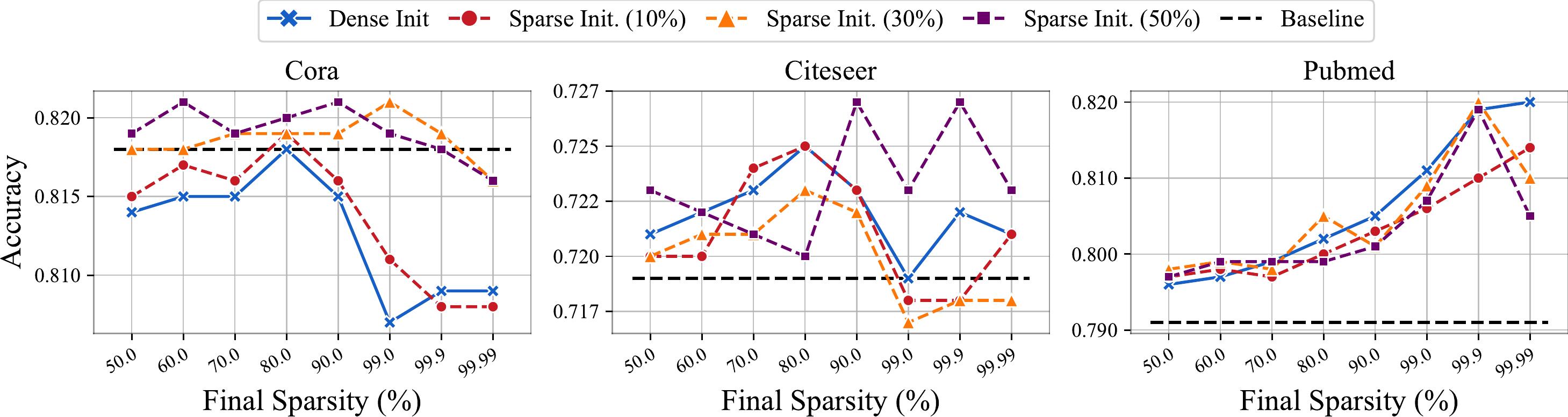}
\end{center}
\caption{Sparse-to-sparse training. More results are given in Appendix~\ref{sec:apd-exp-sts}.}
\label{fig:main-sts}
\end{figure}

\subsection{Further Study: Sparse-to-Sparse Training Schema}

Different from LTH-based methods, which only perform pruning on dense networks and full graphs, our proposed CGP can also support a sparse-to-sparse training schema, which means that our method can be trained with sparse initialization weights and graphs. In this section, we take sparsifying weights as an example and illustrate the results in Figure~\ref{fig:main-sts}. Note that except for the initial density of model weights, other experimental settings are the same as those in the above experiments. Somewhat surprisingly but fully justifiable, it seems that the smaller the initial sparsity is (up to 50\%), the higher the test accuracy of CGP will be.  In particular, on Cora and Citeseer datasets, 50\% weight sparsity initialization usually outperforms the dense initialization, which provides a possibility to train large or deep GNN models on smaller-capacity devices. We leave the training with a sparse initial graph structure and node features for future work.

\section{Conclusion}

 To boost both training and inference efficiency, we have proposed an effective during-training graph pruning (CGP) framework to dynamically prune GNNs from three aspects ($\boldsymbol{A}$,  $\boldsymbol{W}$, and $\boldsymbol{X}$) at the training stage and within one single training process, which helps greatly save the computational cost and achieve comparable or even better performance compared with the LTH-based methods. We have evaluated the proposed method over the node classification task across 6 (various) GNN architectures, on a total of 14 real-world graph datasets. The experiment results have demonstrated that our proposed method boosts both higher training and inference efficiency while offering comparable or even higher accuracy as compared to the state-of-the-art methods. Furthermore, our method contributes to the reduction of both energy costs and $\text{CO}_2$ emissions caused by the training of machine learning models.
 
In future work, \textbf{1)} we will further investigate channel pruning, including the input node features pruning and the hidden embeddings pruning. Also, \textbf{2)} we are planning to design some effective sparse graph structures and node feature initializations, which will guarantee the training of large or deep models on large-scale graph datasets with a small-sized memory CPU/GPU. Furthermore, \textbf{3)} we will work on how to regrow the edges from the complete edge space, not only from the existing edges. In addition, \textbf{4)} we will generalize our method to other graph tasks, such as graph classification and link prediction, to boost their efficiency and accuracy.


\bibliographystyle{IEEEtran}
\bibliography{reference}



\newpage

\appendices

\section{Benchmark Datasets}
\label{sec:apd-dataset}

\begin{table*}[!h]
\caption{Statistics and properties of benchmark datasets.}
\label{tab:dataset}
\resizebox{\textwidth}{!}{%
\begin{tabular}{@{}lllllccl@{}}
\toprule
\multicolumn{1}{c}{}            & \multicolumn{1}{c}{\textbf{Dataset}} & \multicolumn{1}{c}{\textbf{ \# Nodes}} & \multicolumn{1}{c}{\textbf{\# Edges}} & \multicolumn{1}{c}{\textbf{\# Fea.}} & \textbf{\# Cla.} & \textbf{Homo.} & \textbf{Degree} \\ \midrule
\multirow{3}{*}{Small Citation} & Cora                        & 2,708                        & 5,429                        & 1,433                           & 7          & 0.83 & $3.90_{\pm5.23}$   \\
                                & CiteSeer                    & 3,327                        & 4,732                        & 3,703                           & 6          & 0.72 & $2.78_{\pm3.39}$   \\
                                & PubMed                      & 19,717                       & 44,338                       & 500                             & 3          & 0.79 & $4.50_{\pm7.43}$    \\ \midrule
\multirow{6}{*}{Small Hetero}   & Cornell                     & 183                          & 280                          &   1703                              & 5          & 0.30 &  $1.83_{\pm1.71}$    \\
                                & Texas                       & 183                          & 195                          & 1703                                & 5          & 0.11 & $1.83_{\pm1.71}$   \\
                                & Wisconsin                   & 251                          & 466                          &  1703                               & 5          & 0.21 & $1.83_{\pm1.71}$   \\
                                & Actor                       & 7,600                        & 26,752                       &   932                              & 5          & 0.22 & $1.83_{\pm1.71}$  \\\midrule
\multirow{5}{*}{Medium}         & Physics                     & 34,493                       & 247,962                      & 500                             & 5          & 0.91 & $14.38_{\pm15.57}$     \\
                                & CS                          & 18,333                       & 81,894                       & 500                             & 15         & 0.83  & $8.93_{\pm9.11}$   \\
                                & Photo                       & 7,650                        & 119,081                      & 745                             & 8          & 0.85  & $31.13_{\pm47.27}$   \\
                                & Computers                   & 13,752                       & 245,861                      & 767                             & 10         & 0.81 & $35.76_{\pm70.30}$   \\
                                \midrule
\multirow{3}{*}{OGBN}           & Arxiv                       & 169,343                      & 1,166,343                    & 128                             & 40         & 0.63 & $7.68_{\pm9.05}$    \\
                                & Proteins                    & 132,534                      & 39,561,252                   & 8                               & 2          & -- &$597.00_{\pm621.48}$  \\
                                & Products                    & 2,449,029                    & 61,859,140                   &    100                             &   47         & 0.81 & $50.51_{\pm95.91}$    \\ \bottomrule
\end{tabular}%
}
\end{table*}

\subsection{Small-scale Datasets}

We use a total of 7 small-scale datasets involving citation networks (Cora, CiteSeer, and PubMed), web page networks (Cornell, Texas, and Wisconsin), and film-director-actor-writer network (Actor).

\textbf{Cora, CiteSeer, PubMed}  These three benchmark datasets citation networks, which are the most commonly used in the semi-supervised node classification task with the transductive setting.  We strictly follow the train/validation/test split of previous work~\cite{gcn} (i.e., 20 samples per class for training, 500 samples for the validation, and 1000 samples for the test)

\textbf{Cornell, Texas, Wisconsin} These three datasets are web page networks, drawn from computer science departments of three universities (Cornell, Texas, and Wisconsin), where nodes represent web pages and edges represent hyperlinks between them. The task is to classify the nodes into one of the five categories: course, faculty, student, project, and staff. We follow the dataset split of previous work~\cite{h2gcn}.

\textbf{Actor} This dataset is a graph representing actor co-occurrence, drawn from  the film-director-actor-writer network, where nodes represent actors, and edges  denote the author co-occurrence on the same Wikipedia page. The task is to classify the nodes into five categories. We also follow the dataset split of previous work~\cite{h2gcn}.

\subsection{Middle-scale Datasets}

We use a total of 4 middle-scale datasets involving co-author networks and co-purchase networks.

\textbf{Photo, Computers} These two datasets are co-purchase graphs drawn from Amazon, where nodes represent goods and edges represent that two goods are frequently bought together. The task is to classify the categories of goods. We follow the dataset split of previous work~\cite{pitfall}.

\textbf{CS, Physics}  These two datasets are co-author networks, where nodes represent authors and edges denote the co-authored relations of two authors. The task is to classify each author’s respective field of study. We follow the dataset split of previous work~\cite{pitfall}.

\subsection{Large-scale Datasets (OGB)}

To further show the effectiveness and efficiency of our models, we also use 3 large-scale datasets~\cite{ogb-dataset}.

\textbf{ogbn-arxiv} The ogbn-arxiv is a citation network, where nodes represent arXiv papers and edges represent the citations between papers. The task is to classify the areas of arXiv papers. We follow the split provided by~\cite{ogb-dataset}.

\textbf{ogbn-proteins} The ogbn-proteins dataset is protein graph, where nodes represent proteins and edges represent different types of biologically meaningful associations between proteins. Note that, the edges are along with 8-dimensional features. The task is to predict the presence of protein functions in a multi-label binary classification task and the performance is measured by the average of ROC-AUC scores across total 112 tasks. We follow the split provided by~\cite{ogb-dataset}.

\textbf{ogbn-products} The ogbn-products dataset is a product co-purchasing graph, where nodes represent products and edges indicate that the products are purchased together. The task is to classify the category of the products. We follow ~\cite{ogb-dataset} to split the datasets.

\section{Models and Methods}
\label{sec:apd-model}

\subsection{Baseline Models}

\textbf{1) GCN~\cite{gcn}} This method is a baseline model for semi-supervised node classification.

\textbf{2) GAT~\cite{gat}} This method leverage the attention mechanism to assign corresponding weights to their neighbors.

\textbf{3) SGC~\cite{sgcn}} This method removes the nonlinearities in graph convolutional methods, which significantly reduces the training time while maintaining the performance. Specifically, SGC adopts the $K$-th power of the graph convolution matrix in a single neural network layer, which helps to capture higher-order information in the graph.

\textbf{4) APPNP~\cite{appnp} } This method incorporates the Personalized PageRank~\cite{pagerank} into the propagation scheme, which helps to increase the range.

\subsection{SOTA Sparsification Methods for GNNs}

\textbf{5) UGS~\cite{ugs}} This method extends the \textit{lottery ticket hypothesis (LTH)}~\cite{lth} to GNNs and propose the Graph Lottery Ticket (GLT) method to co-sparsify the input graphs and model parameters. 

\textbf{6) GEBT~\cite{gebt} } This method demonstrates that the Early-Bird Tickets Hypothesis~\cite{lth-early-bird} still holds for GNNs, and further developed efficient and effective detectors to automatically identify them, boosting both training and inference efficiency of GNNs.

\textbf{7) ICPG~\cite{icpg}} This method proposes the AutoMasker strategy to endow the GLT with the inductive pruning capacity.

\subsection{SOTA Heterophilous GNN Methods}

 \textbf{8) LINK~\cite{logistic}} This method (logistic regression on the adjacency matrix) only uses the graph topology to perform predictions.
 
 \textbf{9) MLP~\cite{deep-learning}} This method (multilayer perceptron) only leverages the node features to make predictions.  
 
 \textbf{10) H2GCN~\cite{h2gcn}} This method attempts to boost learning from the graph structure under heterophily with the help of a set of key designs, including ego- and neighbor-embedding separation, higher-order neighborhoods, and a combination of intermediate representations.
 
 \textbf{11) MixHop~\cite{mixhop}} This method proposes a graph convolutional layer that utilizes multiple powers of the adjacency matrix, which allows a model to learn general mixing of neighborhood information.
 
 \textbf{12) GPRGNN~\cite{gprgnn}} This method leverages the Generalized PageRank (GPR) to jointly optimize node feature and topological information extraction.

 \textbf{13) FAGCN~\cite{fagcn}} This method is a frequency adaptation graph convolutional network used to adaptively combine the low-frequency and high-frequency signals with a self-gating mechanism.

\section{More Experimental Settings}
\label{sec:apd-exp-setting}

\begin{table*}[!t]
\caption{Implementation details}
\label{tab:apd-hyper}
\renewcommand{\arraystretch}{1.5}
\resizebox{\textwidth}{!}{%
\begin{tabular}{llccccl}
\hline
\multirow{2}{*}{ \textbf{Model}}                                          & \multicolumn{1}{c}{\multirow{2}{*}{ \textbf{Dataset}}}                                                                      & \multicolumn{5}{c}{ \textbf{Hyper-parameter}}                                                                                                                                                                                                                                                              \\ \cline{3-7} 
                                                                & \multicolumn{1}{c}{}                                                                                              & Epochs & \begin{tabular}[c]{@{}c@{}}Learning \\ Rate\end{tabular} & \begin{tabular}[c]{@{}c@{}}Weight \\ Decay\end{tabular} & \begin{tabular}[c]{@{}c@{}}Hidden \\ Units\end{tabular} & \multicolumn{1}{c}{Other}                                                                                \\ \hline
\begin{tabular}[c]{@{}l@{}}GCN, GAT, \\ SGC, APPNP\end{tabular} & All small- and middle-scale datasets                                                                              & 200    & 1e-2                                                     & 5e-4                                                    & 512                                                     & --                                                                                                       \\ \hline
\multirow{2}{*}{UGS}                                            & Cora                                                                                                              & 200    & 8e-3                                                     & 8e-5                                                    & 512                                                     & \multirow{2}{*}{\begin{tabular}[c]{@{}l@{}}Graph and model \\ sparsity regularizers : 1e-2\end{tabular}} \\
                                                                & \begin{tabular}[c]{@{}l@{}}Citeseer, PubMed, \\ Cornell, Texas, Wisconsin,\\ Computers, Photots\end{tabular}      & 200    & 1e-2                                                     & 5e-4                                                    & 512                                                     &                                                                                                          \\ \hline
ICPG                                                            & \begin{tabular}[c]{@{}l@{}}Cora, Citeseer, PubMed, \\ Cornell, Texas, Wisconsin,\\ Computers, Photos\end{tabular} & 200    & 1e-2                                                     & 5e-4                                                    & 512                                                     & \begin{tabular}[c]{@{}l@{}}AutoMasker hidden units: 128;\\ AutMasker learning rate: 1e-2\end{tabular}    \\ \hline
GEBT                                                            & \begin{tabular}[c]{@{}l@{}}Cora, Citeseer, \\ Cornell, Texas, Wisconsin,\\ Computers, Photos\end{tabular}         & 100    & 1e-2                                                     & 5e-4                                                    & 512                                                     & \begin{tabular}[c]{@{}l@{}}Iteration Pruning;\\ EB-tickect from 20 epochs.\end{tabular}                  \\ \hline
\multirow{4}{*}{CGP (Our)}                                      & All small- and middle-scale datasets                                                                              & 200    & 1e-2                                                     & 5e-4                                                    & 512                                                     & \multirow{4}{*}{--}                                                                                      \\
                                                                & ogbn-arxiv                                                                                                        & 500    & 1e-2                                                     & 0                                                       & 64                                                      &                                                                                                          \\
                                                                & ogbn-proteins                                                                                                     & 200    & 1e-2                                                     & 0                                                       & 64                                                      &                                                                                                          \\
                                                                & ogbn-products                                                                                                     & 100    & 1e-3                                                     & 0                                                       & 128                                                     &                                                                                                          \\ \hline
\end{tabular}%
}
\end{table*}

\textbf{Baseline Models} Since we experiment with a large number of models and datasets, we maintain almost the same configurations across all models and datasets. Specifically, for the four baseline models, GCN~\cite{gcn}, GAT~\cite{gat}, SGC~\cite{sgcn}, and APPNP~\cite{appnp}, we train all models with $200$ epochs (training iterations) using Adam~\cite{adam} with a learning rate of $0.01$ and a weight decay of $5\mathrm{e}-4$ (See Table~\ref{tab:apd-hyper}).  The hidden units of each model are fixed at $512$, except GAT which is set at $8$.  The dropout rate of all models is fixed at $0.5$. We select the best epoch to test according to the accuracy of the validation datasets. Additionally, for GAT~\cite{gat}, we use 8 heads in the first layer, and 1 head in the second layer. For SGC~\cite{sgcn},  the hop is set as $2$. For APPNP, the number of iterations is set as $10$, and the teleport probability is set as $0.1$.

\textbf{Our Proposed CGP} For a fair comparison, we adopt the same settings as the ones of the above baseline models (See Tabel~\ref{tab:apd-hyper}). Additionally, for all models and datasets, we tune three hyper-parameters we introduced by grid search. We choose regrowth rate $r$ from $\{0.1, 0.2, 0.3\}$, pruning frequency $\Delta t $ from $\{10, 20, 30\}$, and final pruning epoch from $\{50, 100, 150\}$.

\textbf{SOTA Sparsification Methods for GNNs} We strictly follow the settings provided in the original papers. The detailed implementations are described in Table~\ref{tab:apd-hyper}.

\textbf{SOTA Hetero-GNN Methods}  We mainly follow the settings adopted in a recent benchmarking work~\cite{benchmark-heter} for non-homophilous graph datasets. All methods are optimized by adam~\cite{adam}, with test performance reported for the learned parameters of highest validation performance. For all methods, we tune the hidden units from $\{8, 16, 32, 64, 128, 256, 512\}$, the dropout rate from $\{0.0,  0.5\}$ on $4$ heterophilous datasets (Cornell, Texas, Wisconsin, and Actor). In addition, 
\begin{itemize}
	\item H2GCN~\cite{h2gcn}: we train for 200 epochs.
	\item  MixHop~\cite{mixhop}: we train for 400 epochs. We tune the number of MixHop Graph Convolution layers from $\{1, 2, 3\}$, and each layer uses the 0th, 1st, and 2nd powers of the adjacency with ReLU activations. The last layer is a linear projection layer, instead of the attention output mechanism in~\cite{mixhop}.
	\item GPRGNN~\cite{gprgnn}: we train for 400 epochs. We tune the learning rate from $\{0.01, 0.05, 0.002\}$, the a polynomial graph filter of order from $\{5, 10, 15\}$, and the teleport probability from $\{0.1, 0.2, 0.5, 0.9\}$.
	\item FAGCN~\cite{fagcn}: we train for 200 epochs.  We additionally tune the number of FAconv layer from $\{1, 2\}$, and the scaling hyper-parameter $\epsilon$ from $\{0.1, 0.2, 0.3\}$.
\end{itemize}

\section{More Related Works}
\label{sec:apd-related-work}

\textbf{Graph Sampling} Graph sampling is an intuitive solution to tackle the problem of huge computation cost. There are mainly two types of graph sampling methods, neighbor sampling and subgraph sampling. Neighbor sampling~\cite{graphsage,pinsage,vr-gcn,fastgcn} selects a fixed number of neighbors for each node, while subgraph sampling~\cite{cluster-gcn,graphsaint,rwt,pgs} samples a set of subgraphs in the training process. The former ensures that every node can be sampled but may suffer from  the problem of neighbor explosion, causing the  exponential increase of  both the training and inference time. The latter cannot guarantee that every node can be at least sampled once in the whole training/inference process. Thus it is only feasible for the training process, because the testing process usually requires GNNs to process each node in the graph. For a more detailed description, please refer to~\cite{acceleration-survey,survey-sampling}.

\textbf{Graph Condensation} Graph condensation~\cite{graph-condense} also aims to obtain a smaller size graph. However, graph condensation does not reduce the number of nodes as done in graph coarsening works~\cite{graph-coarsen-cut, graph-coarsen-scale}, but aims to learn synthetic nodes and connections with the help of the gradient matching method~\cite{data-condense}. Specifically,  it matches the network parameters w.r.t. large-real and small-synthetic training data by matching their gradients at each training step. In this way, the training trajectory on a small-synthetic graph can mimic that of the original large-real graph.

\textbf{Graph Pooling}  Graph pooling is also an effective approach to generate a smaller sized graph. The methods of graph pooling can be roughly divided into \textbf{flat pooling} and \textbf{hierarchical pooling}. The former directly generates a graph-level representation in one step, mostly taking the average or sum over all node embeddings as the graph representation~\cite{duvenaud}, while the latter coarsens the graph gradually into a smaller size graph, mainly in two manners: \textbf{node clustering pooling}~\cite{diffpool,mincut,structpool,mem-pool} and \textbf{node drop pooling}~\cite{graph-u-net,sagpool,gsapool,hgp-sl,TAPool,vip-pool,MVpool,NDPool}. Specifically, node clustering pooling generates new nodes to construct the coarsened graph by arranging nodes into clusters, which is time-and space-consuming~\cite{mincut}. In contrast, node drop pooling selects a subset of nodes from the original graph to construct the coarsened graph, which is more efficient and more fit for large-scale graphs~\cite{sagpool} but suffers from inevitable information loss~\cite{ipool}. For a more detailed description, please refer to the recent proposed review of graph pooling~\cite{pooling-survey}. However, all the above methods are primarily designed to obtain a higher accuracy in the graph classification task and the smaller graphs are a byproduct obtained during the training process.

\textbf{Graph Structure Learning} Similar to the graph structure pruning method we proposed in Section~\ref{sec:method}, graph structure learning also attempts to learn an optimized graph structure for downstream tasks by cutting off or adding, since they think the original noisy or incomplete graphs often lead to unsatisfactory representations~\cite{agcn,prognn,idgl,slaps,vib-gsl}. For a more detailed description, please refer to~\cite{structure-learning-survey}. However, these methods are incapable of learning graphs with a smaller size, and are thus not applicable for sparse graph training.

\section{More Experiment Results}
\label{sec:apd-exp-results}

\subsection{Proposed CGP vs. Baseline Models}
\label{sec:apd-vs-base}

\begin{figure*}[!ht] 
\begin{center}
\includegraphics[width=1.0\linewidth]{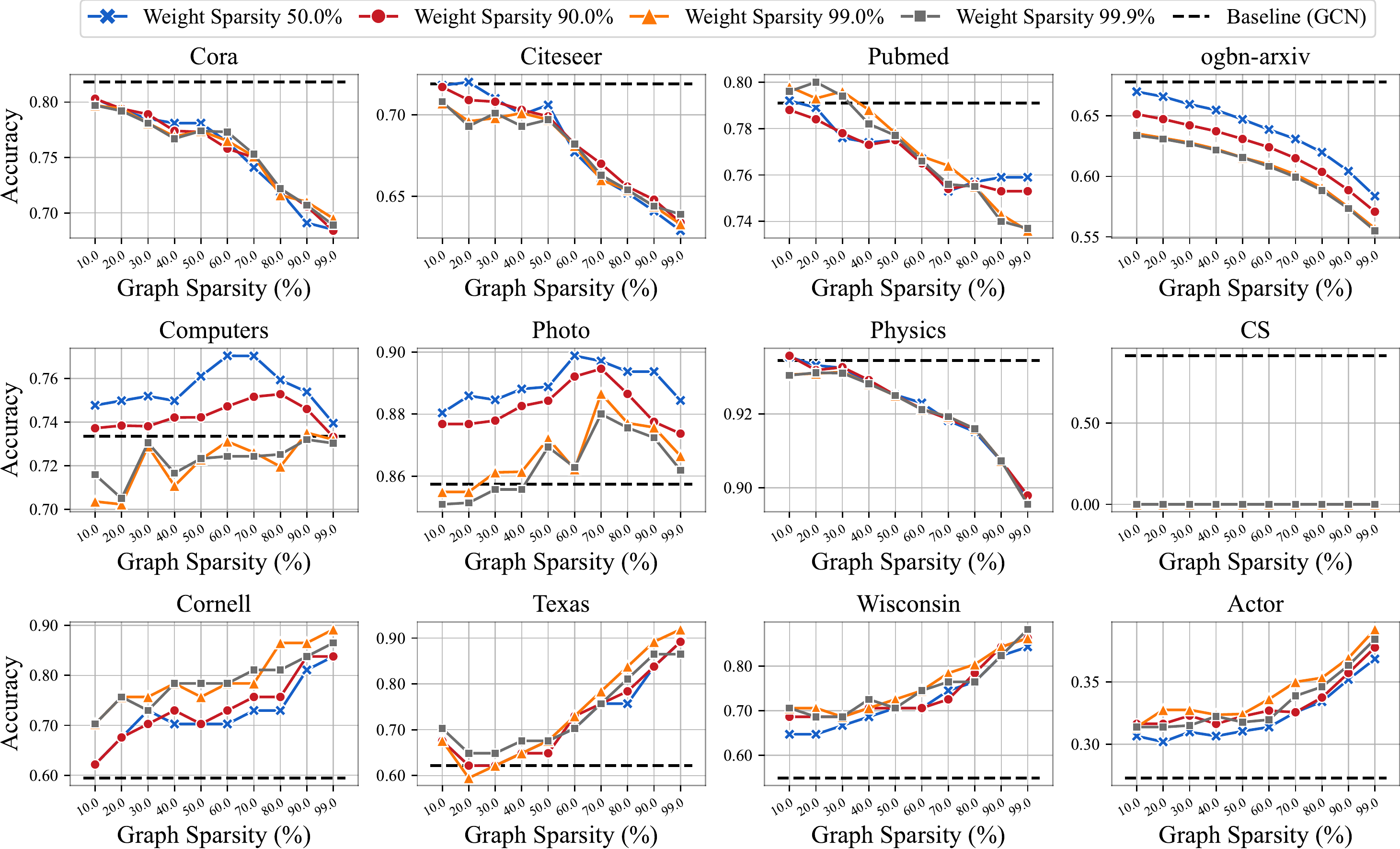}
\end{center}
\caption{Co-Sparsifying model weighs and graph structure (GCN). }
\label{fig:apd-wa-gcn}
\end{figure*}

\begin{figure*}[!ht] 
\begin{center}
\includegraphics[width=1.0\linewidth]{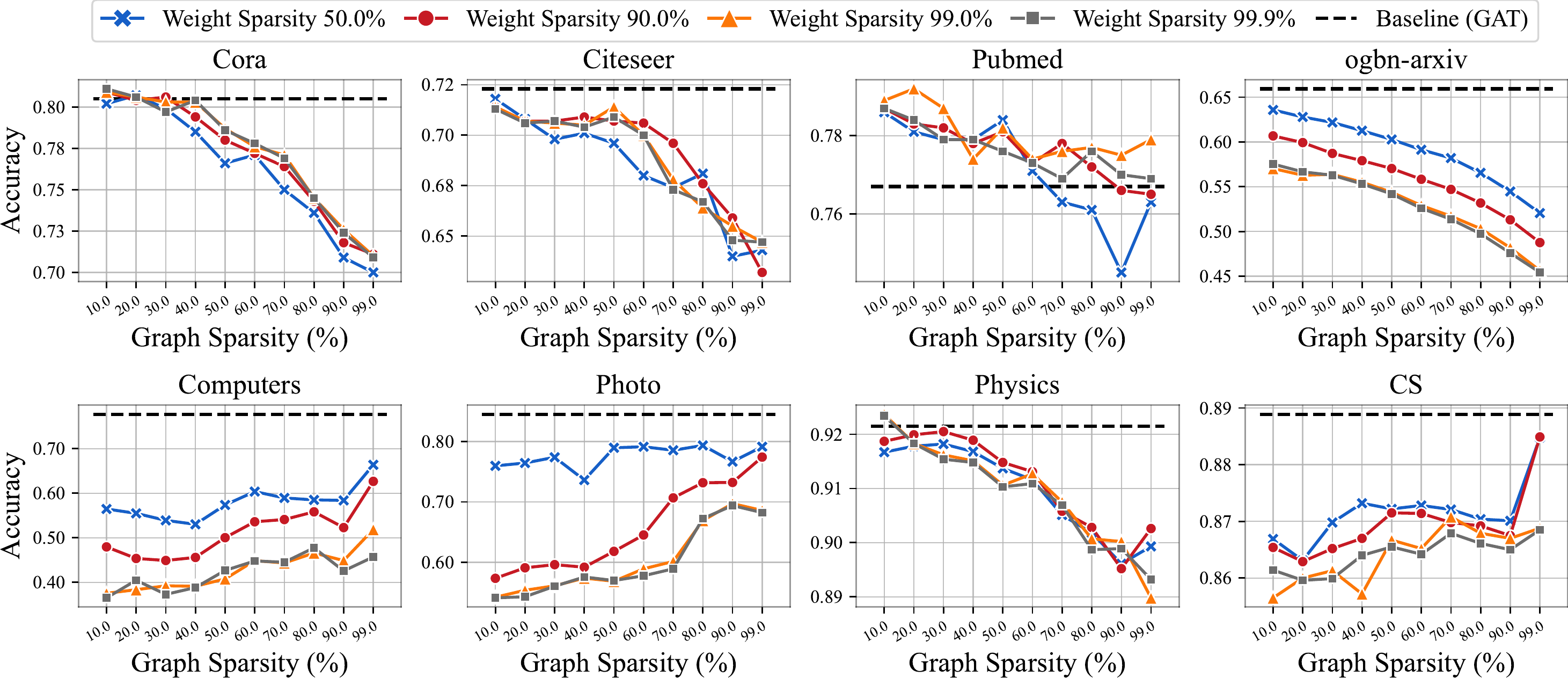}
\end{center}
\caption{Co-Sparsifying model weighs and graph structure (GAT). }
\label{fig:apd-wa-gat}
\end{figure*}

\begin{figure*}[!ht] 
\begin{center}
\includegraphics[width=1.0\linewidth]{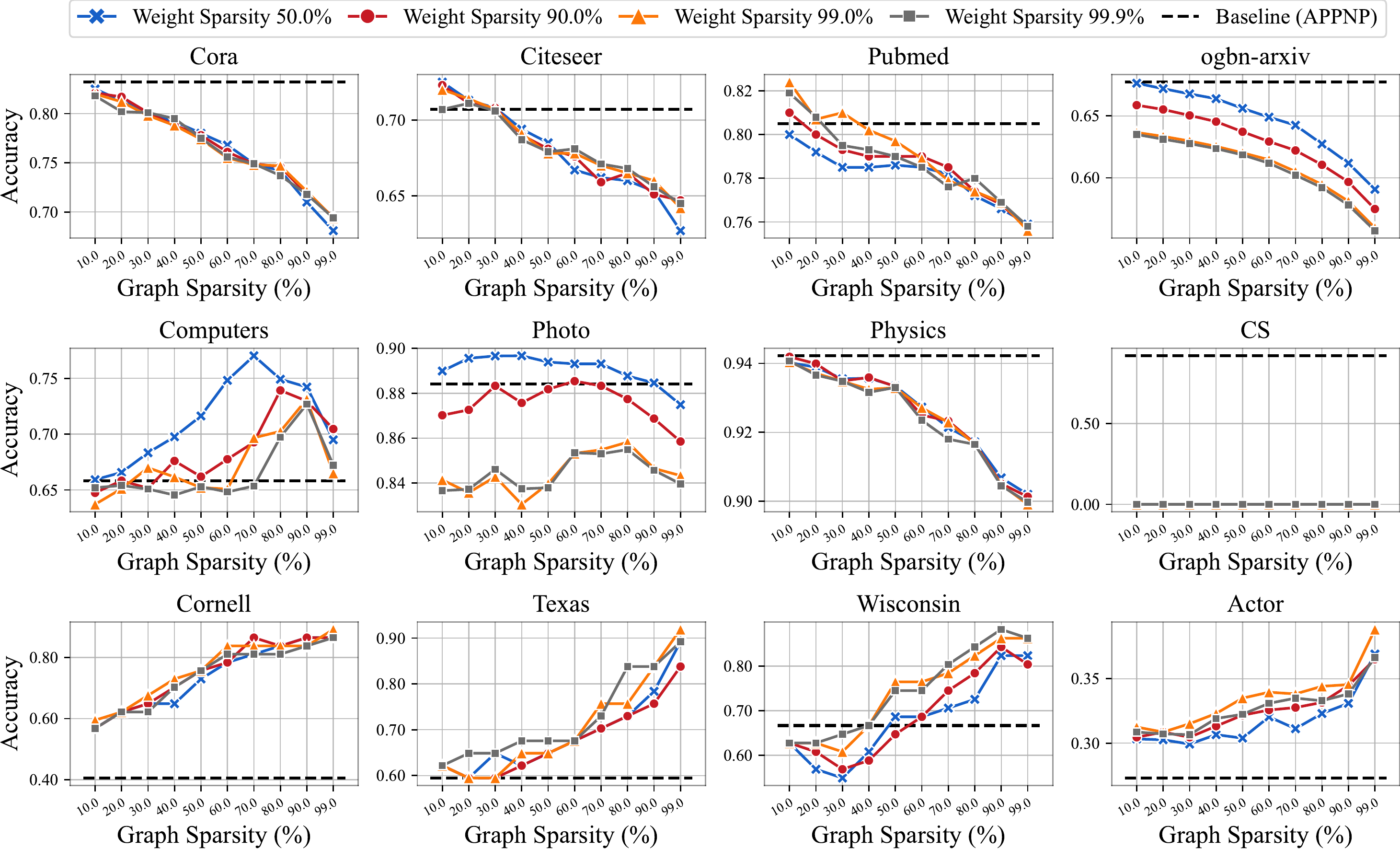}
\end{center}
\caption{Co-Sparsifying model weighs and graph structure (APPNP). }
\label{fig:apd-wa-appnp}
\end{figure*}

\begin{figure*}[!ht] 
\begin{center}
\includegraphics[width=1.0\linewidth]{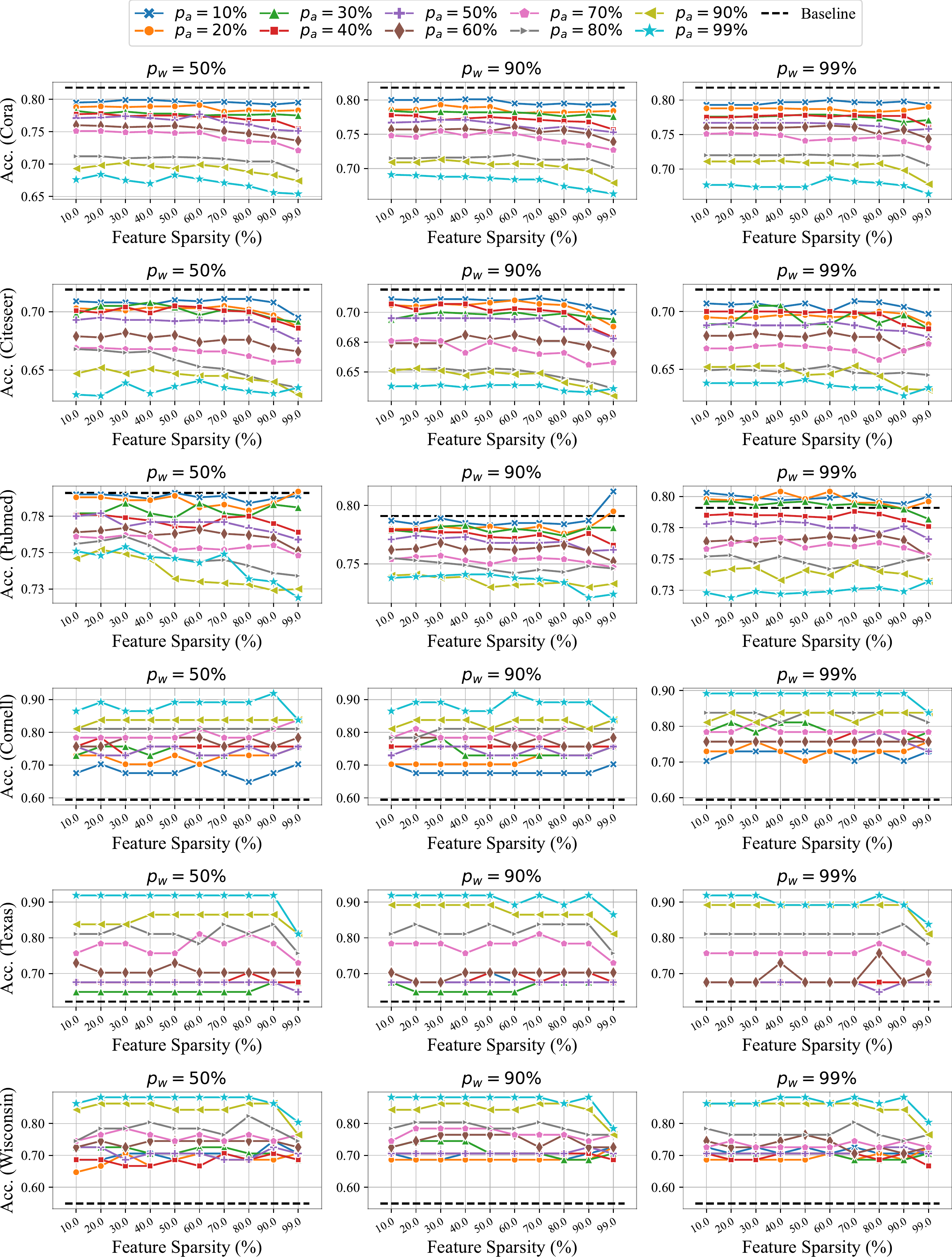}
\end{center}
\caption{Co-Sparsifying model weighs, graph structure, and node features (GCN). }
\label{fig:apd-waf}
\end{figure*}

\begin{figure*}[!ht] 
\begin{center}
\includegraphics[width=1.0\linewidth]{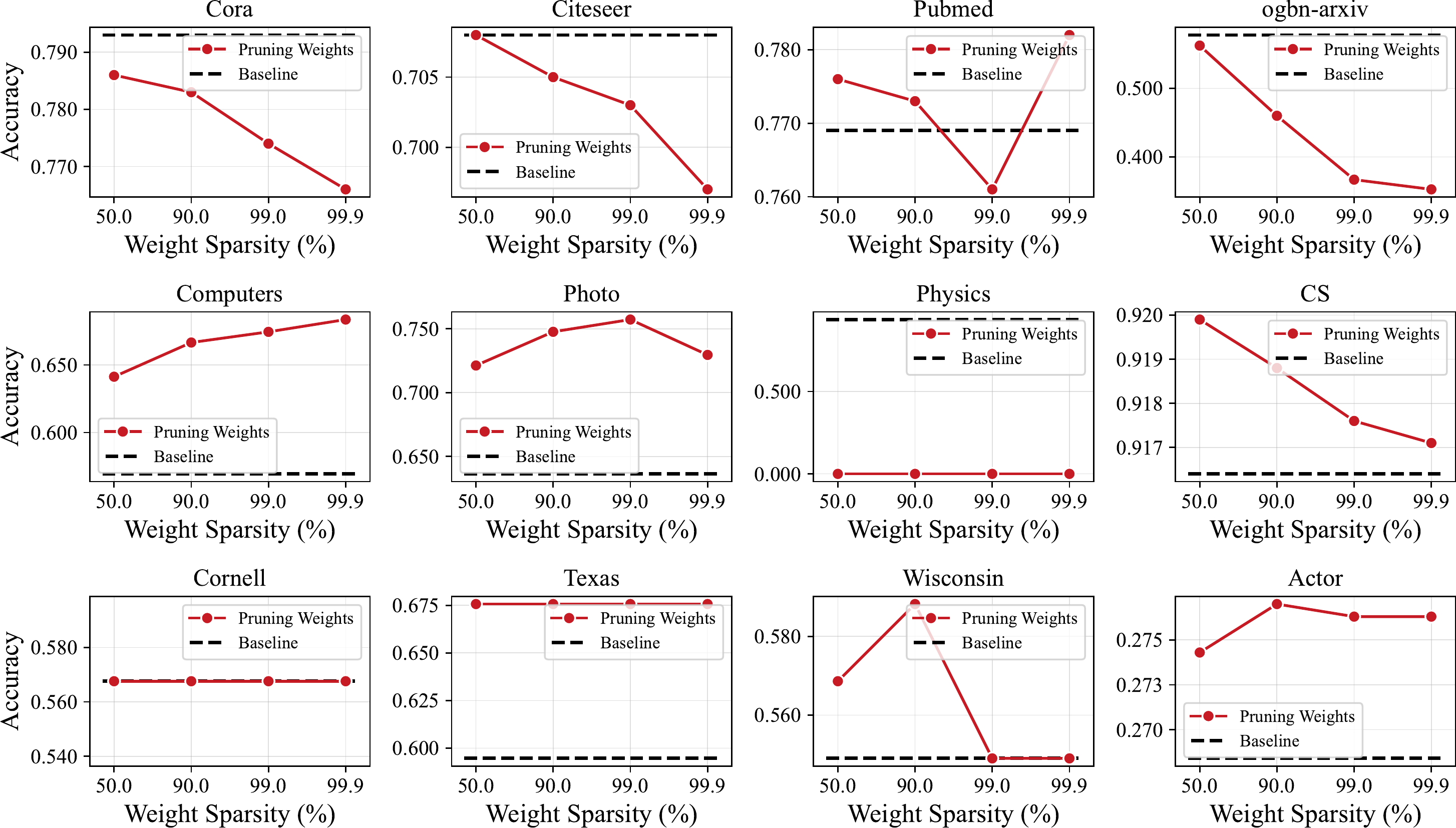}
\end{center}
\caption{Sparsifying model weighs on SGC model over 12 graph datasets. }
\label{fig:apd-w-sgc}
\end{figure*}

The results of co-sparsifying model weights and graph structure over GCN, GAT, and APPNP are shown in Figures~\ref{fig:apd-wa-gcn},~\ref{fig:apd-wa-gat}, and~\ref{fig:apd-wa-appnp}, respectively. The results of co-sparsifying model weights, graph structure, and node features over GCN are shown in Figure~\ref{fig:apd-waf}. The results of sparsifying model weights over SGC are shown in Figure~\ref{fig:apd-w-sgc}.

\subsection{Proposed CGP Vs. SOTA Sparsification Methods for GNNs.}
\label{sec:apd-exp-sota-gnn}

\begin{figure*}[!ht] 
\begin{center}
\includegraphics[width=0.9\linewidth]{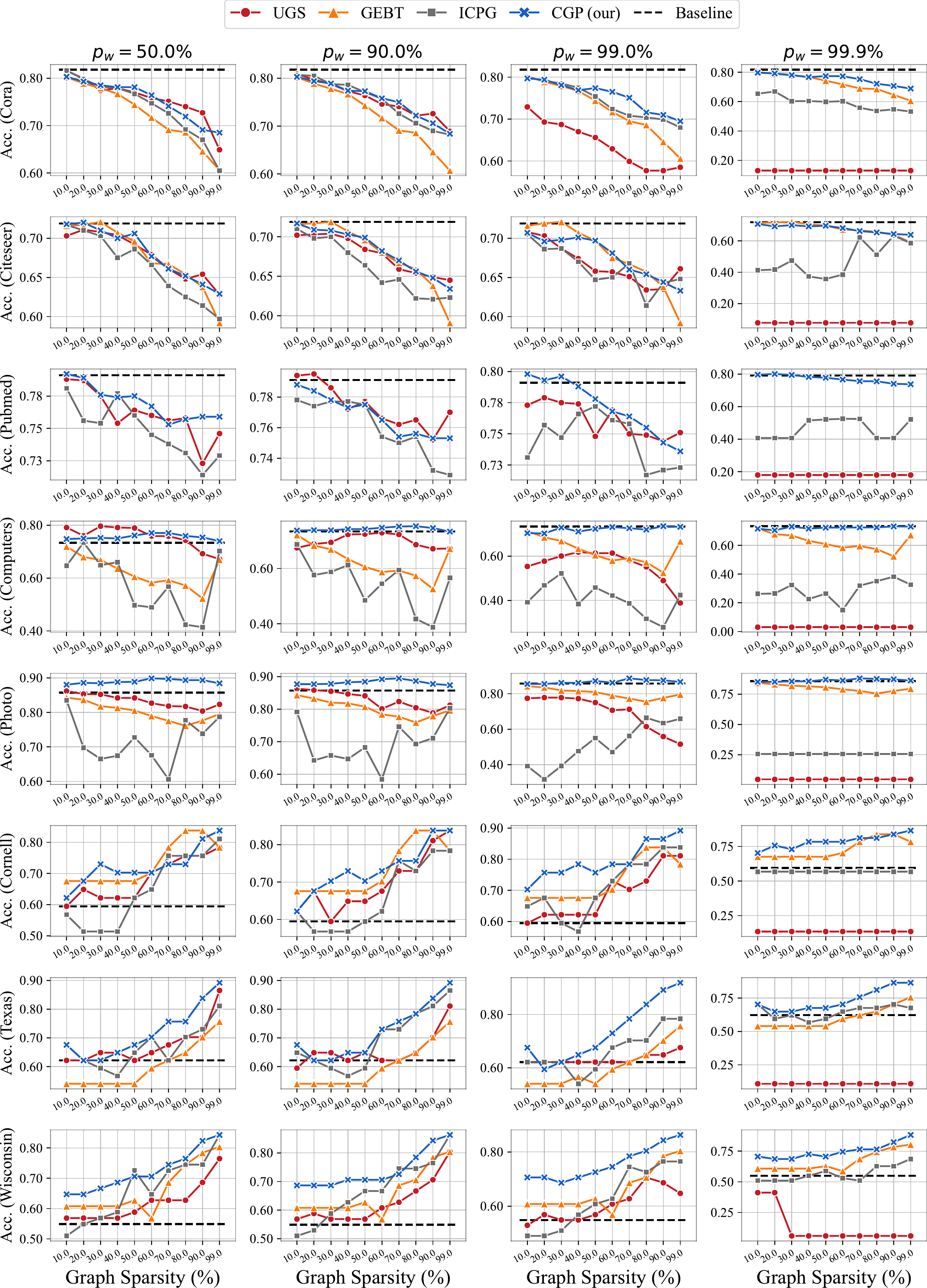}
\end{center}
\caption{Comparing with sparsification methods for GNNs. }
\label{fig:apd-vs-sota}
\end{figure*}

\begin{figure*}[!ht] 
\begin{center}
\includegraphics[width=1.0\linewidth]{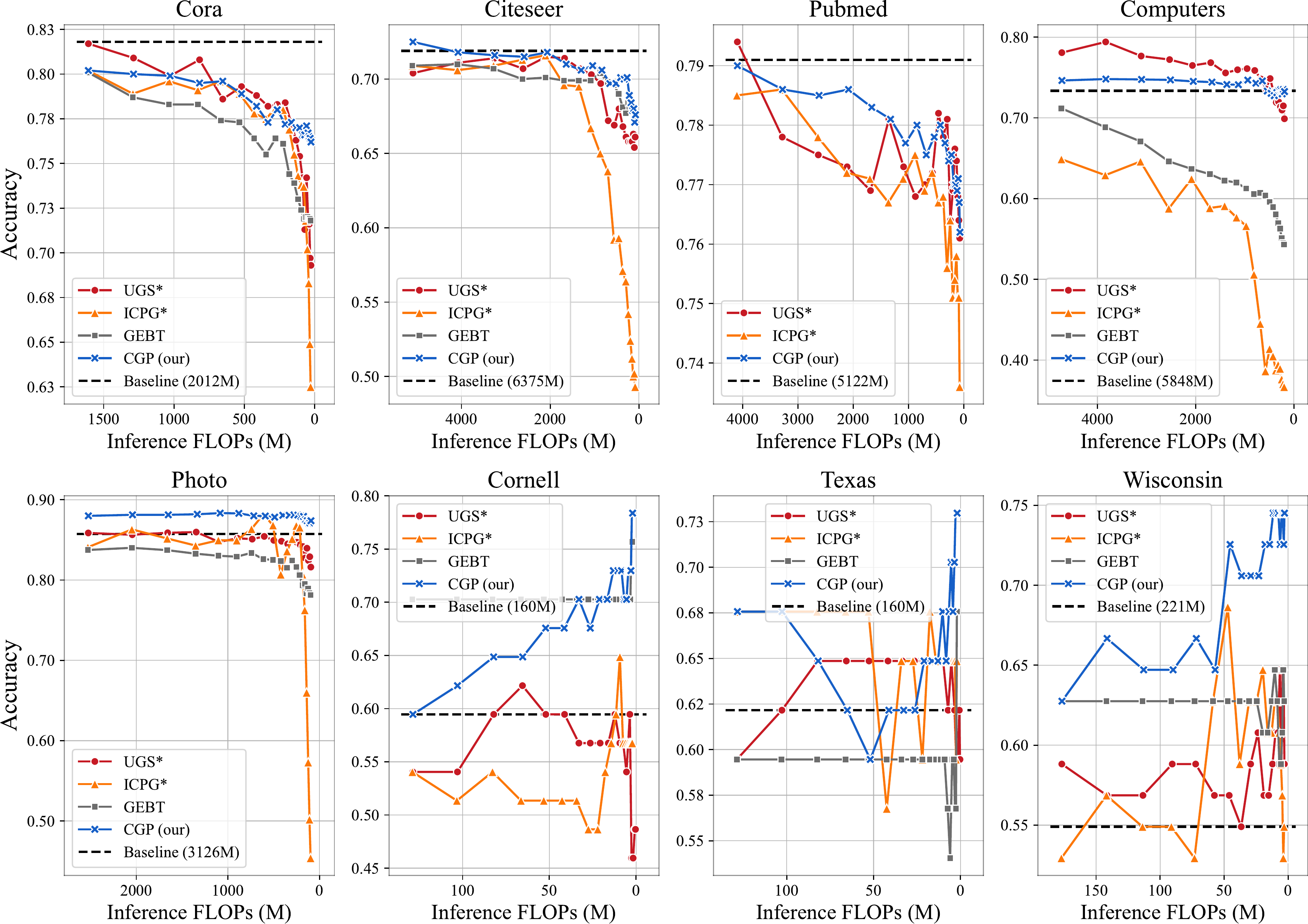}
\end{center}
\caption{Comparing with sparsification methods for GNNs in the context of FLOPs. }
\label{fig:apd-flops}
\end{figure*}

More results on the comparison with SOTA sparsification methods are shown in Figures~\ref{fig:apd-vs-sota} and~\ref{fig:apd-flops}. From the results, we can observe that our CGP method consistently performs better than other SOTA methods in most situations.

\subsection{Further Proposed CGP on SOTA Heterophilous GNN Models}
\label{sec:apd-exp-sota-hetero}

\begin{figure*}[!ht] 
\begin{center}
\includegraphics[width=1.0\linewidth]{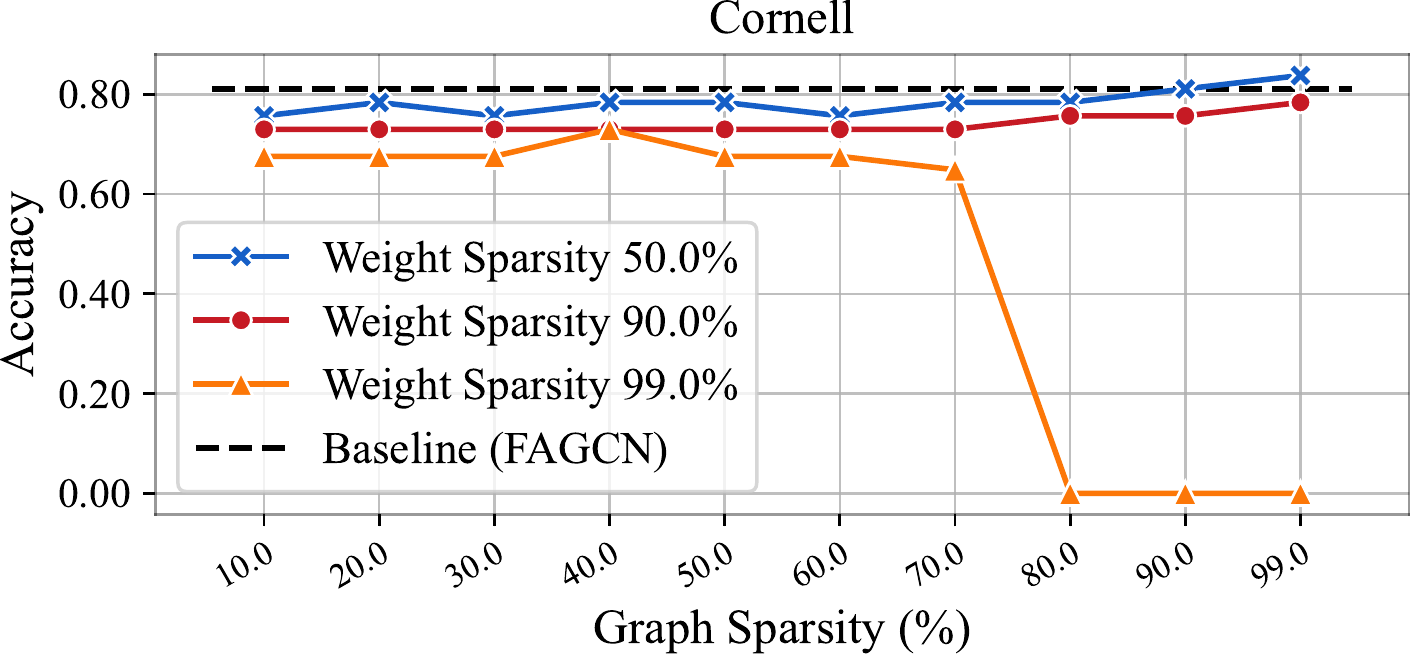}
\end{center}
\caption{ Applying our method on SOTA heterophilous GNNs (FAGCN). }
\label{fig:apd-heter-fagcn}
\end{figure*}

Furthermore, we apply our method on FAGCN~\cite{fagcn} to explore whether our CGP method can further improve the effectiveness of the SOTA Heterophilous GNN models. As shown in Figure~\ref{fig:apd-heter-fagcn}, when the graph sparsity becomes larger, our method can provide a comparable performance while lowering the inference cost.

\subsection{Sparse to Sparse Training Schema}
\label{sec:apd-exp-sts}

\begin{figure*}[!ht] 
\begin{center}
\includegraphics[width=1.0\linewidth]{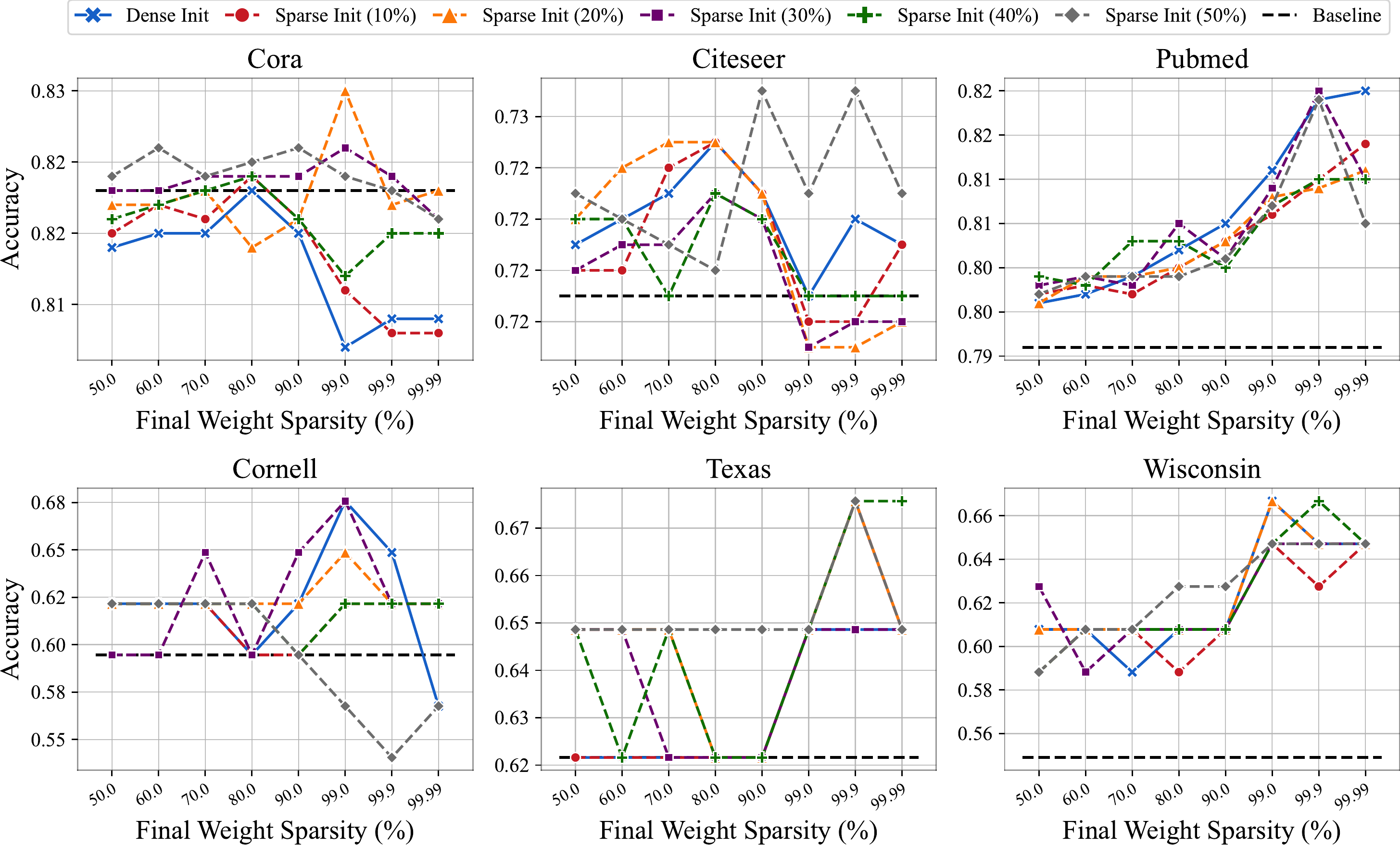}
\end{center}
\caption{Sparse-to-Sparse training schema. }
\label{fig:apd-sts}
\end{figure*}

In this section, we demonstrate more results of the sparse-to-sparse training schema over three heterophilous datasets, shown in Figure~\ref{fig:apd-sts}. 

\subsection{Ablation Study}
\label{sec:apd-exp-ablation}

\begin{figure*}[!h] 
\begin{center}
\includegraphics[width=0.9\linewidth]{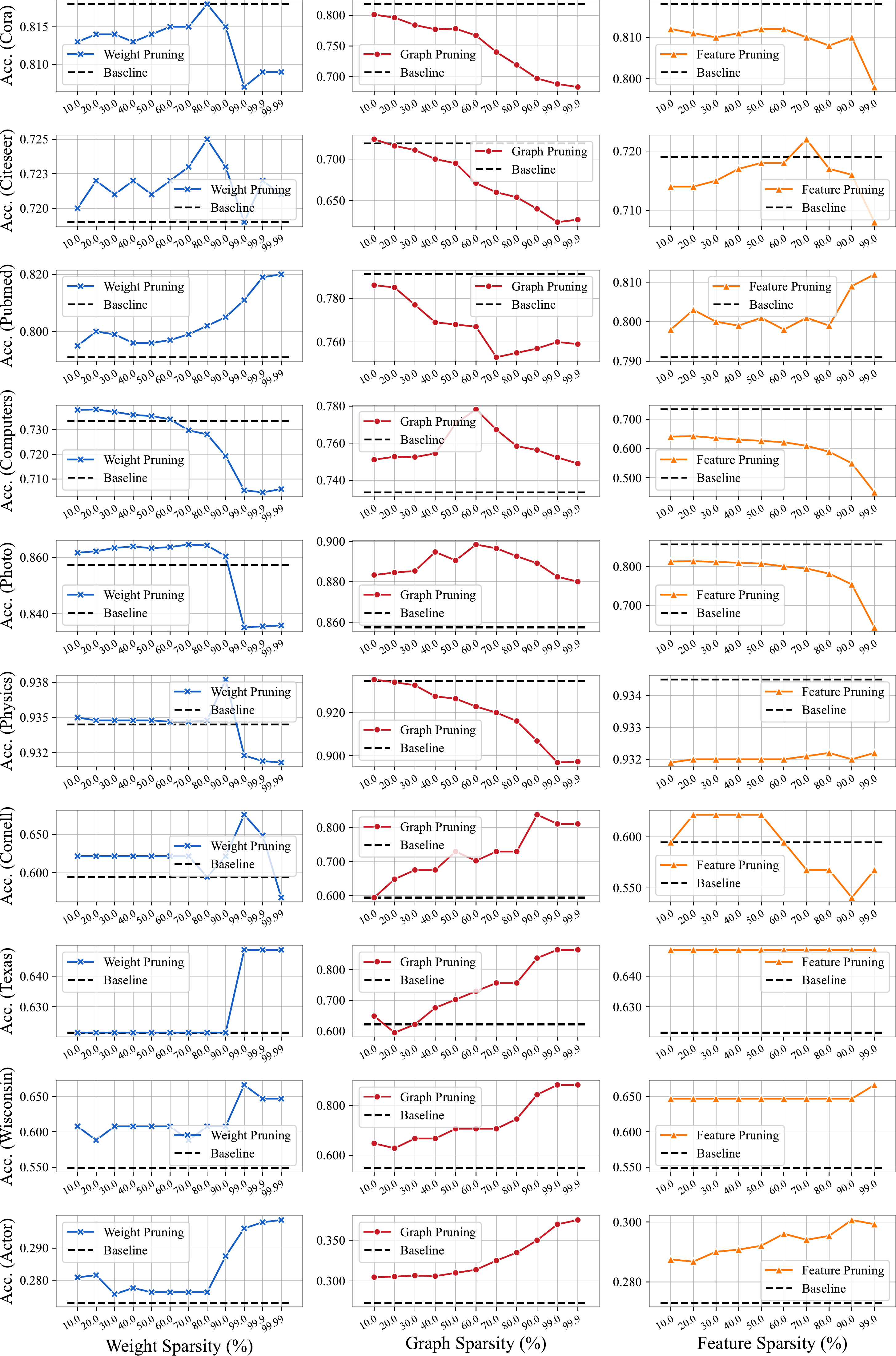}
\end{center}
\caption{Ablation Study. }
\label{fig:apd-ablation}
\end{figure*}

\begin{figure*}[!h] 
\begin{center}
\includegraphics[width=1.0\linewidth]{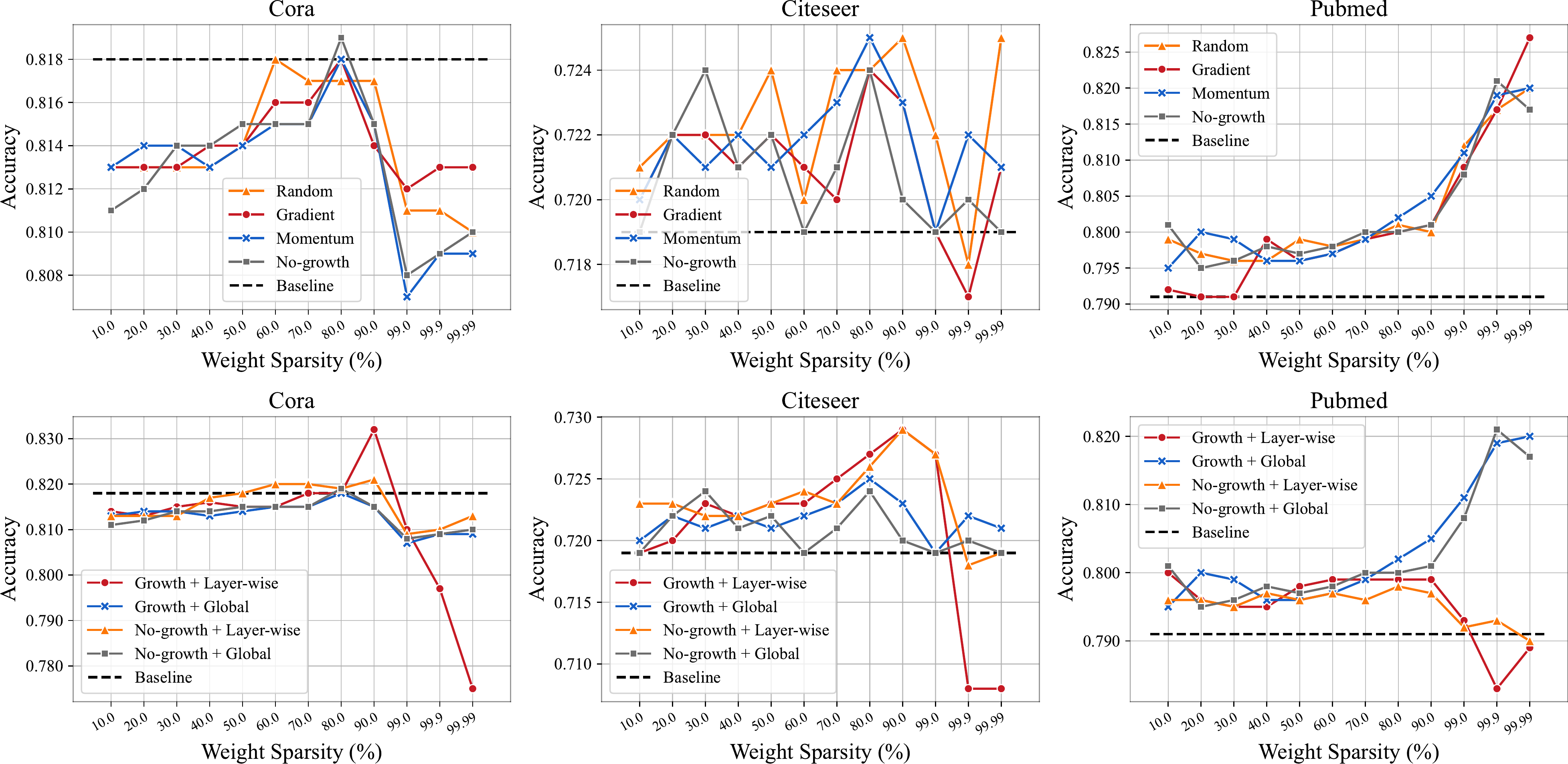}
\end{center}
\caption{Effectiveness of Regrowth and Pruning Way. }
\label{fig:apd-effect}
\end{figure*}

In Figure~\ref{fig:apd-ablation}, we present the results of sparsifying elements separately on GCN over ten graph datasets. We can observe that \textbf{1)} the trend of pruning weights is similar to the one of pruning node features, since pruning features is in fact pruning the channel of weights.  \textbf{2)} Only pruning weights can usually bring an improvement.  \textbf{3)} The node features are redundant in several datasets.  \textbf{4)} Pruning features performs worse in Computers, Photos, and Physics datasets, since the original dimensions of node features in these datasets are small.  Figure~\ref{fig:apd-effect} shows the effectiveness of different regrowth and pruning ways.

\subsection{Training Time Analysis}
\label{sec:apd-exp-train-time}

\begin{figure*}[!h] 
\begin{center}
\includegraphics[width=1.0\linewidth]{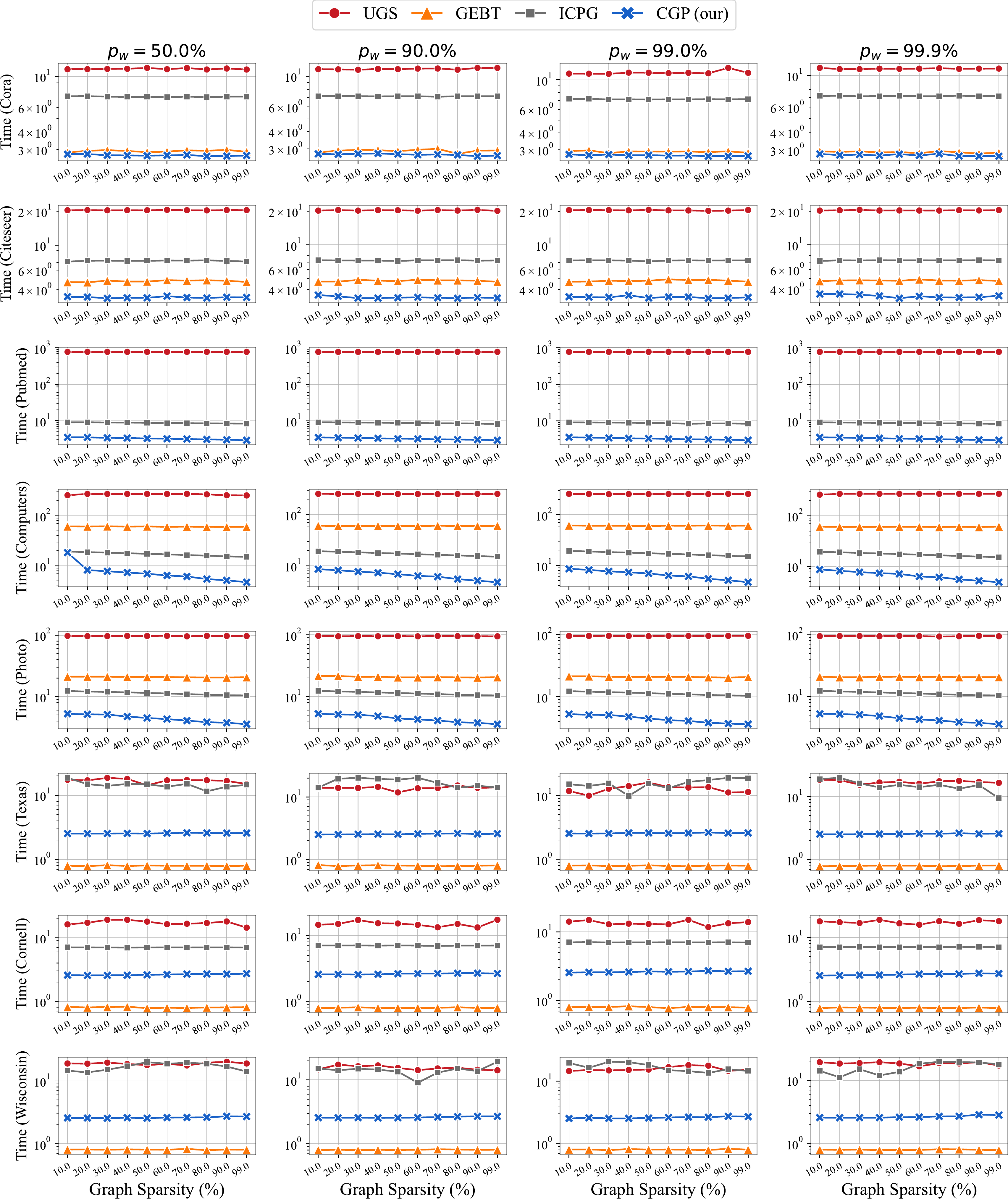}
\end{center}
\caption{Comparison with SOTA LTH-based methods for GNN models in the context of  training time. We co-sparsify model weights and graph structures on GCN model.}
\label{fig:apd-train-time}
\end{figure*}

We compare our CGP with SOTA sparsification methods in the context of training time, shown in Figure~\ref{fig:apd-train-time}. We can observe that \textbf{1)} our CGP is faster in most cases. \textbf{2)} The larger graph datasets become,  the more we CGP has an advantage in training time.  \textbf{3)} Weight sparsity have less influence on the training time.

\subsection{Sensitivity Analysis}
\label{sec:apd-exp-sensitive}

\begin{figure*}[!h] 
\begin{center}
\includegraphics[width=0.9\linewidth]{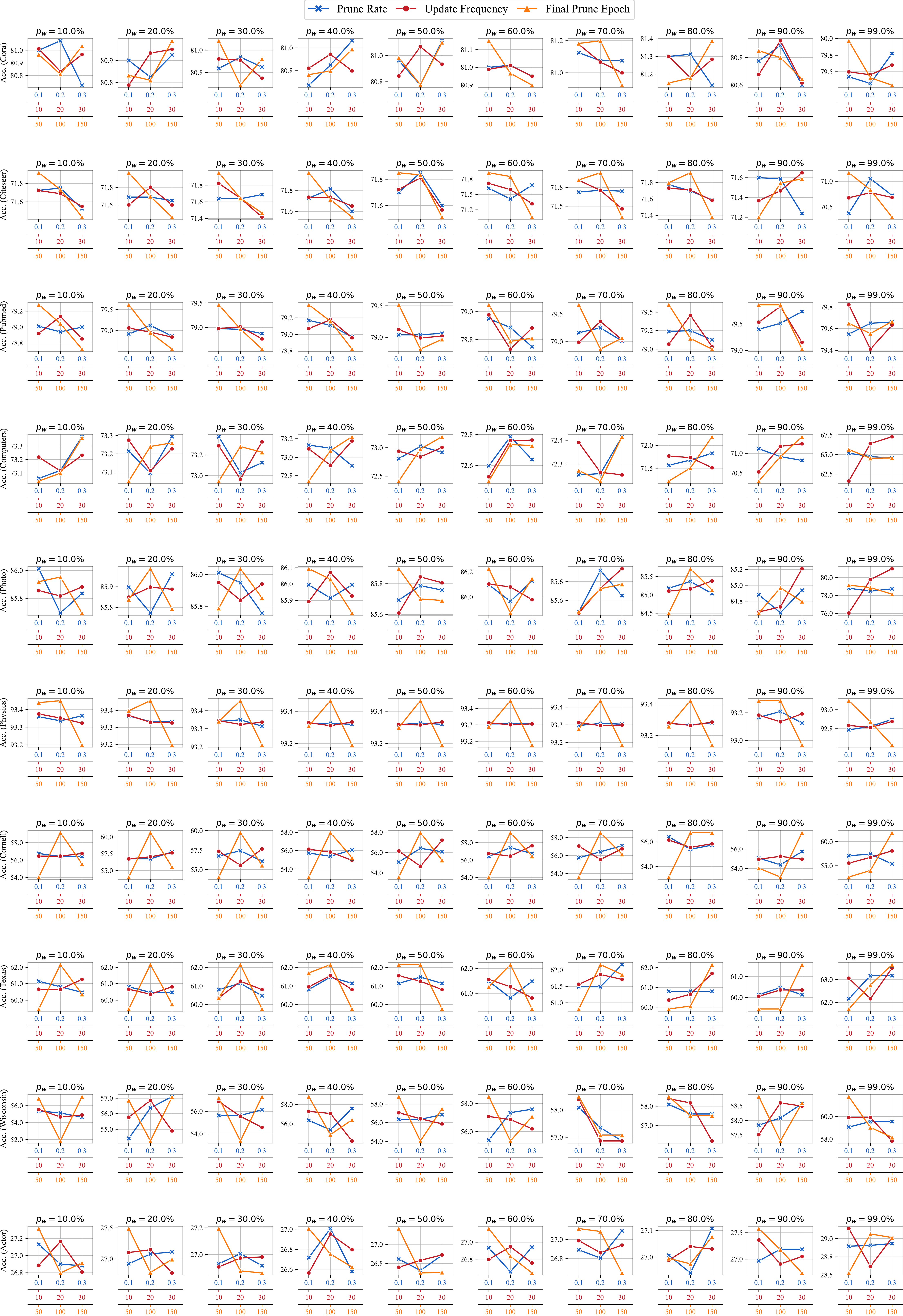}
\end{center}
\caption{Parameter analysis. We sparsify model weights on GCN over 10 graph datasets. }
\label{fig:apd-para-w}
\end{figure*}

\begin{figure*}[!h] 
\begin{center}
\includegraphics[width=0.9\linewidth]{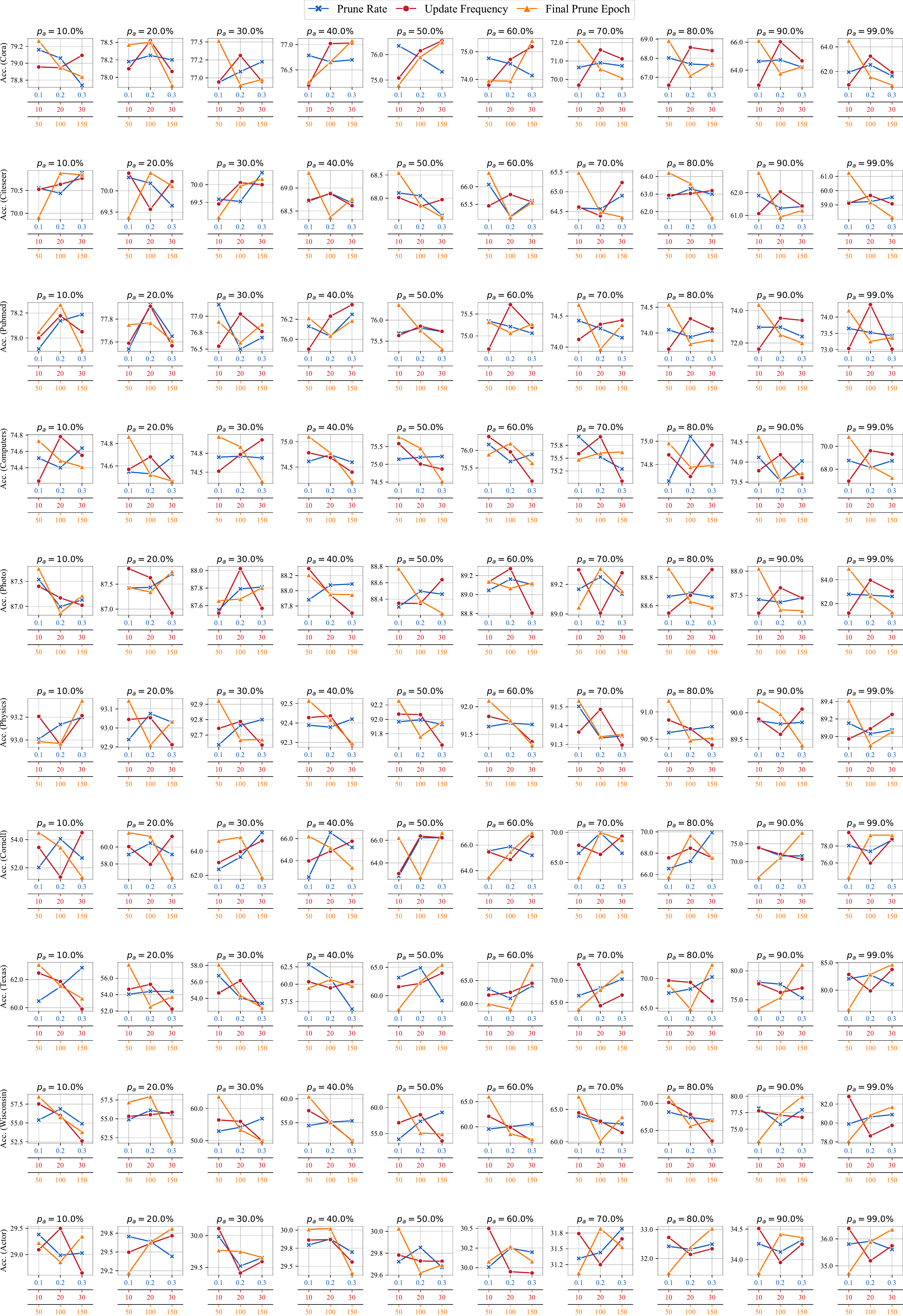}
\end{center}
\caption{Parameter analysis. We sparsify graph structures on GCN over 10 graph datasets. }
\label{fig:apd-para-a}
\end{figure*}

\begin{figure*}[!h] 
\begin{center}
\includegraphics[width=0.9\linewidth]{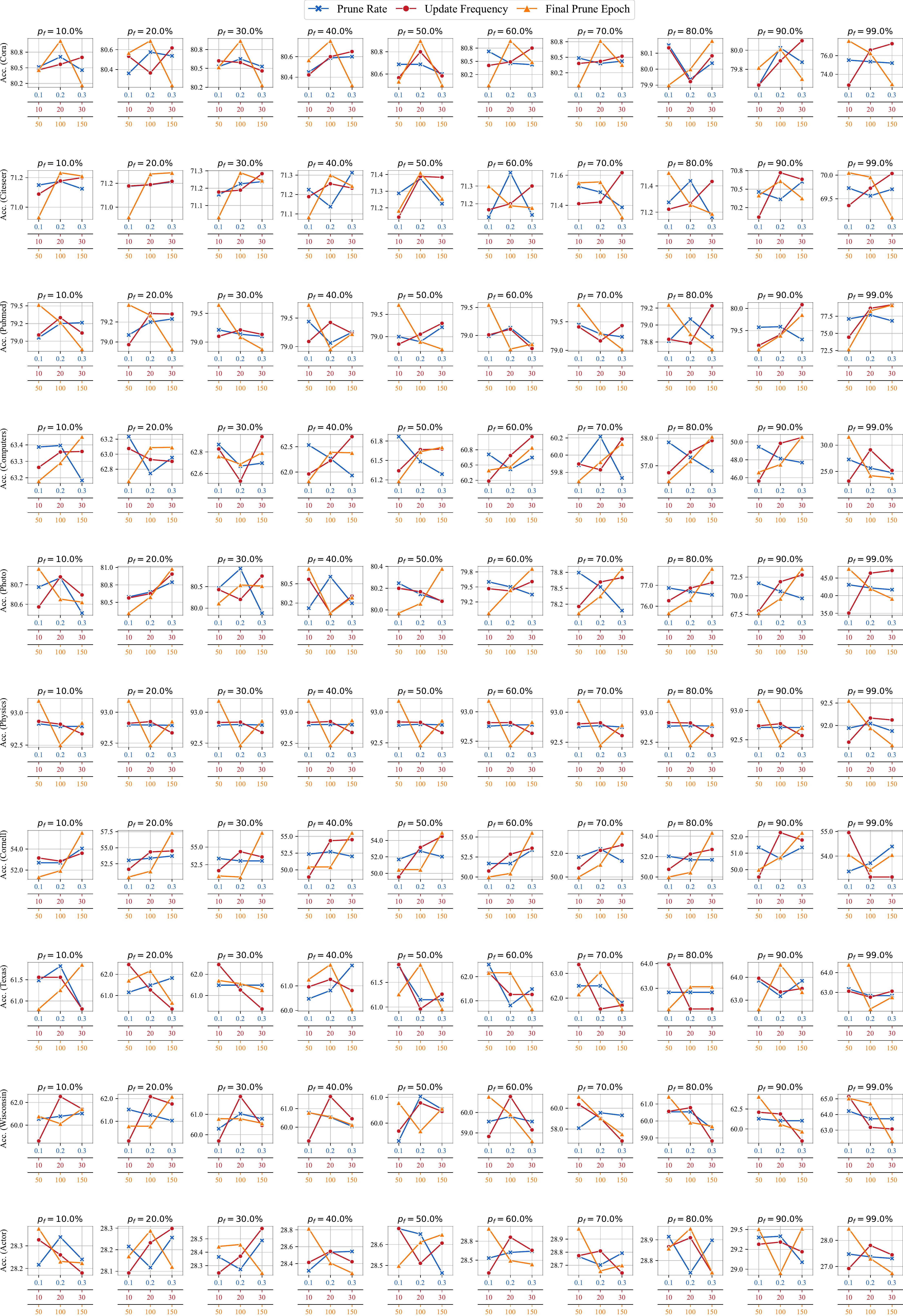}
\end{center}
\caption{Parameter analysis. We sparsify node features on GCN over 10 graph datasets.}
\label{fig:apd-para-x}
\end{figure*}

In this section, we provide a comprehensive analysis of three hyper-parameters: 1) Pruning Rate, 2) Pruning Frequency, 3) Final Epoch. We choose regrowth rate $r$ from $\{0.1, 0.2, 0.3\}$, pruning frequency $\Delta t $ from $\{10, 20, 30\}$, and final pruning epoch from $\{50, 100, 150\}$. We present the results of sparsifying model weights, graph structure, and node features in Figures~\ref{fig:apd-para-w},~\ref{fig:apd-para-a}, and~\ref{fig:apd-para-x}, respectively.

%
%
%
%

\vfill

\end{document}